\documentclass[11pt,a4paper]{article}
\usepackage[hyperref]{eacl2021}
\usepackage{times}
\usepackage{latexsym}

\usepackage[utf8]{inputenc} % allow utf-8 input
\usepackage[T1]{fontenc}    % use 8-bit T1 fonts
\usepackage{hyperref}       % hyperlinks
\usepackage{url}            % simple URL typesetting
\usepackage{booktabs}       % professional-quality tables
\usepackage{amsfonts}       % blackboard math symbols
\usepackage{nicefrac}       % compact symbols for 1/2, etc.
\usepackage{microtype}      % microtypography
\usepackage{dsfont}
\usepackage{multirow}
\usepackage{graphicx}
\usepackage{amsmath}
\usepackage{mathtools}
\usepackage{amssymb}
\usepackage{amsthm}
\usepackage{csquotes}
\usepackage{breqn}
\usepackage{subfig}
\usepackage{caption} 
\usepackage{bbm}
\usepackage{algorithm}
\usepackage[noend]{algpseudocode}
\usepackage{bm}
\usepackage{soul}
\DeclareMathOperator*{\argmax}{arg\,max}

\DeclareMathOperator*{\cossim}{cos\,sim}

\expandafter\patchcmd\csname\string\algorithmic\endcsname%
  {\labelwidth 0.5em}{\labelwidth0pt\labelsep0pt}{}{}

\usepackage{paralist}

\usepackage[normalem]{ulem}
\useunder{\uline}{\ul}{}
\newcommand{\asli}[1]{{\color{green}{\bf\sf [A*]}}}
\newcommand{\dpos}{\ensuremath{\mathop{\mathfrak D}^{+}}\nolimits} 
\newcommand{\dneg}{\ensuremath{\mathop{\mathfrak D}^{-}}\nolimits}

\usepackage[symbol]{footmisc}
\newcommand{\astfootnote}[3]{
\let\oldthefootnote=\thefootnote
\setcounter{footnote}{1}
\renewcommand{\thefootnote}{\fnsymbol{footnote}}
\footnote{#1}
\let\thefootnote=\oldthefootnote
}
\usepackage{microtype}

\aclfinalcopy % Uncomment this line for the final submission
\def\aclpaperid{13} %  Enter the acl Paper ID here

\title{
Contrastive Multi-document Question Generation
}
 
\author{Woon Sang Cho$^{\star\dagger}$ \quad Yizhe Zhang$^{\circ}$\quad Sudha Rao$^{\circ}$\quad Asli Celikyilmaz$^{\circ}$\quad \\
\textbf{Chenyan Xiong$^{\circ}$\quad Jianfeng Gao$^{\circ}$\quad Mengdi Wang$^{\star}$\quad Bill Dolan$^{\circ}$}  \\
  $^{\star}$Princeton University, Princeton, NJ, USA \\
  $^{\circ}$Microsoft Research, Redmond, WA, USA \\
  {\tt $^\star$\{woonsang,mengdiw\}@princeton.edu} \\
  {\tt $^\circ$\{yizzhang,sudhra,aslicel,cxiong,jfgao,billdol\}@microsoft.com}
  }

\date{}
\begin{document}

\maketitle

\begin{abstract}
Multi-document question generation focuses on generating a question that covers the common aspect of multiple documents. Such a model is useful in generating clarifying options.
However, a naive model trained only using the targeted (``positive'') document set may generate too generic questions that cover a larger scope than delineated by the document set.
To address this challenge, we introduce the contrastive learning strategy where given ``positive'' and ``negative'' sets of documents, we generate a question that is closely related to the ``positive'' set but is far away from the ``negative'' set.
This setting allows generated questions to be more specific and related to the target document set.
To generate such specific questions, we propose Multi-Source Coordinated Question Generator (MSCQG), a novel framework that includes a supervised learning (SL) stage and a reinforcement learning (RL) stage. 
In the SL stage, a single-document question generator is trained. 
In the RL stage, a coordinator model is trained to find optimal attention weights to align multiple single-document generators, by optimizing a reward designed to promote specificity of generated questions. 
We also develop an effective auxiliary objective, named Set-induced Contrastive Regularization (SCR) that improves the coordinator's contrastive learning during the RL stage.
We show that our model significantly outperforms several strong baselines, as measured by automatic metrics and human evaluation. 
The source repository is publicly available at \url{www.github.com/woonsangcho/contrast_qgen}.\footnote[2]{Work was done when the author was an intern at Microsoft Research. }
\end{abstract}

\begin{figure}[t!]
\centering
\includegraphics[width=1.0\linewidth]{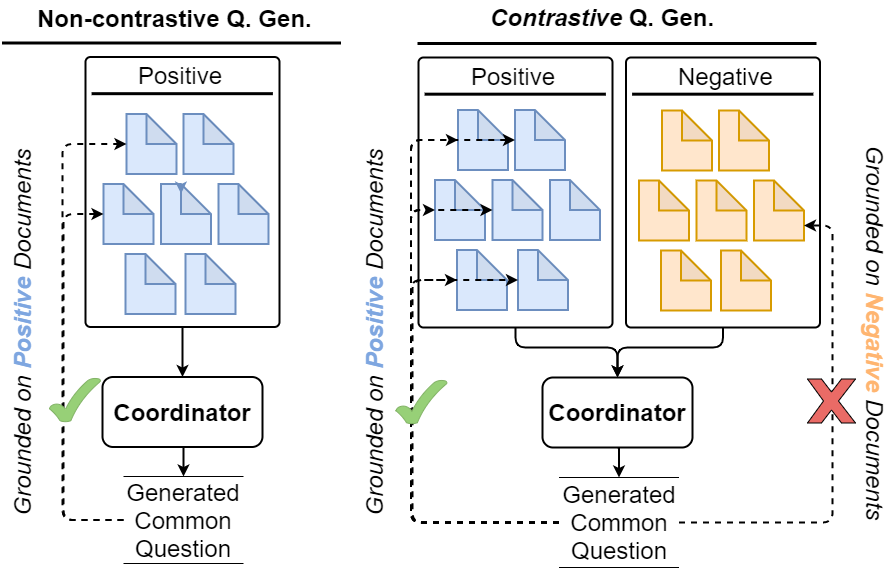}
\resizebox{\linewidth}{!}{%
\begin{tabular}{ll}
\multicolumn{2}{l}{} \\
\multicolumn{1}{l|}{Positive set about:} & \textit{number of saturn's moons} \\
\multicolumn{1}{l|}{Negative set about:} & \textit{uranus how many moons} \\ \hline
\multicolumn{1}{l|}{Non-contrastive Q. Gen} & \textit{what is the largest moon} \\
\multicolumn{1}{l|}{Contrastive Q. Gen} & \textit{how many moons are there in saturn}
\end{tabular}%
}
\caption{Non-contrastive and contrastive method for multidocument question generation. 
Left: non-contrastive modeling that takes input as a set of positive documents.  
However, model-generated questions from this method are rather generic and not specific to the input documents. 
Right: \textit{contrastive} modeling, which considers both positive and negative document sets, and learns to generate questions that are more grounded on the positive document set.}
\label{coordinator}
\vspace{-5mm}
\end{figure}

\section{Introduction}

User queries on web search engines can sometimes be vague. 
%In such a scenario, the document set returned by a search engine could span several different unrelated subtopics making the user search experience difficult and tedious. 
Search engines may resolve this ambiguity by suggesting clarification options back to the user in the form of questions \citep{10.1145/3020165.3022149,10.1145/3331184.3331265,zamani2020generating}.
However, asking the right clarification questions is a challenging information-seeking task, given a plethora of possible questions \citep{rao-daume-iii-2018-learning,rao-daume-iii-2019-answer,qi2020stay}.
One workaround is to take informational cues from the search engine results given the initial query. The clarification options are then generated from non-ranked and non-overlapping thematic partitions of the search engine results.
%, akin to evaluating informativeness through a question answering model \citep{qi2020stay}.
The whole pipeline is akin to the pseudo-relevance feedback \citep{rocchio1971relevance,10.1145/1390334.1390377}.
This can significantly reduce the search space, and has the potential to generate correct clarification questions within the context \citep{cho_qgen_2019}.

This particular approach may involve three non-trivial phases: 
$i)$ \textit{retrieval}: gather the initial return documents by the search engine; $ii)$ \textit{partition}: partition the documents into semantically similar clusters in an unsupervised manner; $iii)$ \textit{multi-document question generation}: generate a clarification question by finding an ``overlap'' among documents in each cluster. 
In principle, the clarification questions should be \textbf{specific} to each cluster rather than generic and bland, otherwise it is counter to the objective of clarification \citep{10.1145/3020165.3020183}.
In this work, we focus on developing a multi-document question generator to generate cluster-specific questions in the $iii)$ step. Nevertheless, we believe our approach can be readily applied to multi-document text generation such as summarization \citep{lapata2019} and response generation \citep{zhang2019dialogpt}.

We address this challenge by leveraging contrastive learning. Given a set of positive documents $\dpos$ and a set of negative documents $\dneg$ (where $\dneg$ is yet semantically close to $\dpos$), we propose a new strategy to generate a question that is semantically relevant to $\dpos$ and far away from $\dneg$. 
Ideally, the model would use both $\dpos$ and $\dneg$ to identify distinguishing features between the two sets and constrain the generation to be \textit{specific} to $\dpos$. 
The similarity between the $\dpos$ and $\dneg$ makes the generation more challenging and forces the model to be as specific as possible in order to distinguish between the two sets.
The comparison between the contrastive and non-contrastive multi-document question generation is illustrated in Figure~\ref{coordinator}.

This task is particularly challenging because $i)$ there does not exist direct supervised \emph{ground-truth} multi-document question given positive and negative sets of documents. 
%$ii)$ The \emph{oracle} human-written queries, detailed in Section~\ref{experiments}, are typically noisy. 
$ii)$ The whole procedure involves multiple aspects including language understanding, inter-document information aggregation, coordinative planning and language generation. 
In theory, the generator can be trained to maximize the chance that the generated question specifically retrieves the given document cluster, using RL. However, the space of possible sequence is prohibitively large which results in large variance in RL \citep{lewis2017deal}. 
To effectively reduce the search space of RL, we employ a hybrid supervised learning (SL) and RL strategy.
% (Section~\ref{model}).
We also propose a novel reward-shaping auxiliary objective, Set-induced Contrastive Regularization (SCR) (Section~\ref{model}), which heuristically drives the generation closer towards $\dpos$, by \emph{minimizing}/\emph{maximizing} the KL divergence between the hypothesis distribution and distributions induced by $\dpos$/$\dneg$.
% This is achieved by \emph{minimizing} the KL divergence between the aggregated word distribution and distributions induced by $\dpos$. Likewise, SCR drives the generation away from $\dneg$ by \emph{maximizing} the KL divergence yet calibrating this effect by monitoring how similar the two sets of distributions induced by $\dpos$ and $\dneg$ are.

Our contributions are summarized below:
% $i)$ We propose a contrastive multi-document question generation task.
$i)$ We develop a novel Multi-Source Coordinated Question Generator (MSCQG) model that is trained using a hybrid hierarchical generation scheme. The document-specific generator is fine-tuned from GPT-2 and the inter-document coordinator is trained using reinforcement learning. 
$ii)$ We introduce Set-induced Contrastive Regularization (SCR), an auxiliary regularizer that pushes MSCQG toward $\dpos$ relative to $\dneg$ while limiting the effect of $\dneg$ in a principled manner. 
$iii)$ Empirical results show that our model is able to generate more grounded and specific questions, significantly outperforming existing baseline models in automatic measures and human evaluation.

\section{Method}\label{model}
\begin{figure}[t!]
\centering
\includegraphics[width=1.0\linewidth]{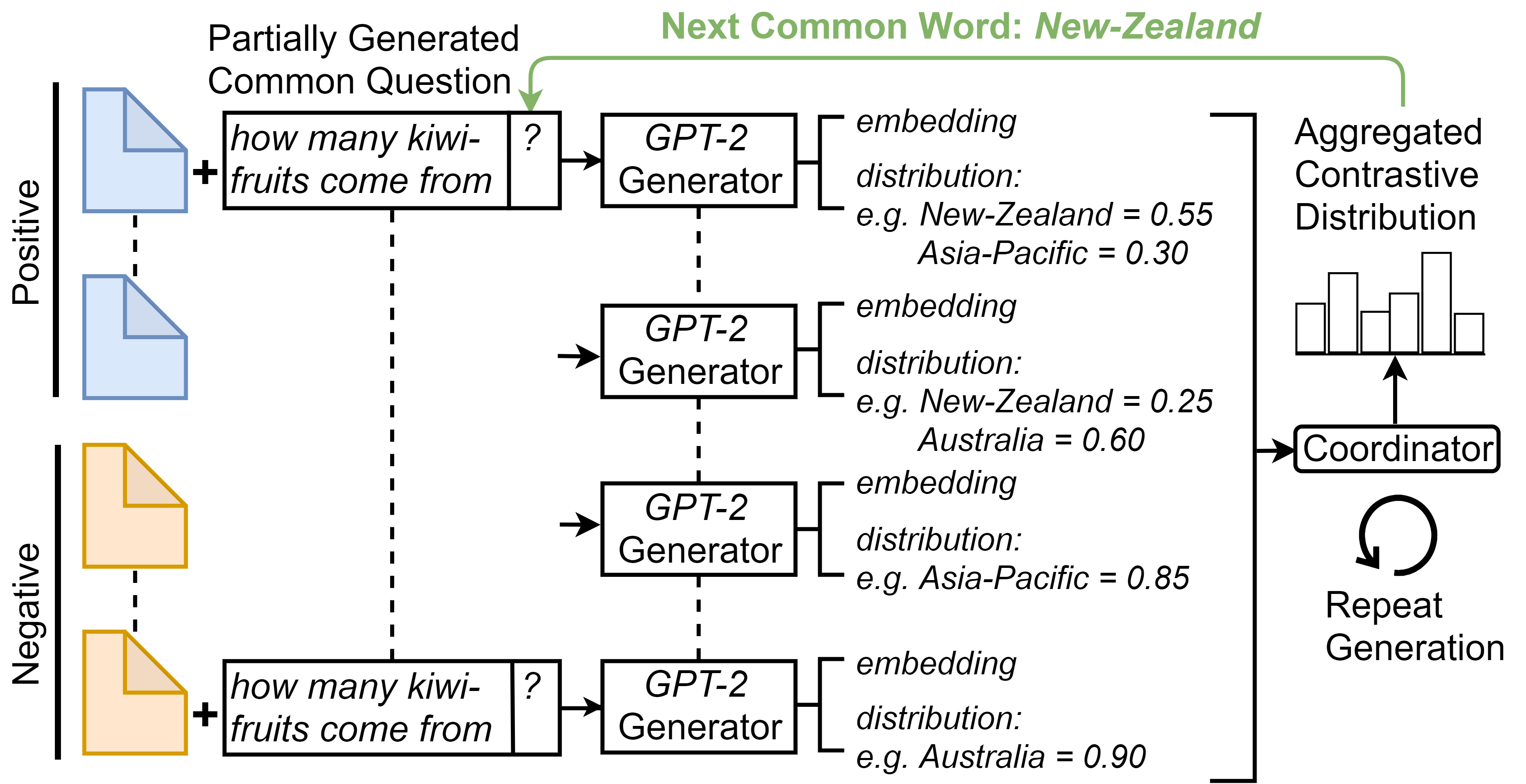}
\caption{System overview. 
The example is an illustration using fictitious tokens for ease of understanding.
Our MSCQG model learns to attend different weights and form a final aggregated distribution at each decoding time.
%, given the input embeddings and distributions.
The decision to enforce or penalize the negative set distributions to the aggregated distribution is controlled in a principled manner. 
% For details, see Section~\ref{model}.
}
\label{system}
\end{figure}

{\setlength{\parindent}{0cm}
\textbf{Overview: }}  The overview of our model is illustrated in Figure~\ref{system}. The model consists of two major components:
$i)$ The \textit{document-specific generator} generates a question from a single document, and is fine-tuned from OpenAI GPT-2.
$ii)$ The \textit{inter-document coordinator} integrates multiple-document information from the \textit{document-specific generator} instances. The coordinator is trained using reinforcement learning after fixing the document-specific generator.

During the RL training, at each generation time step, each (positive and negative) document will independently use the same generator trained from $i)$ to predict the next token. The coordinator will learn to aggregate the probabilities to a \textit{consensus probability} by maximizing a reward function. The reward function is designed to encourage the generated question to tie to the positive set and to be away from the negative set. The word newly generated from the \textit{consensus probability} are concatenated to all documents as inputs for next time step. \newline

% This coordinator is trained using reinforcement learning and an auxiliary regularization to encourage (or discourage) the generation to be more relevant to the positive (or negative) set.
% The source code and dataset will be released upon publication. \newline
% Specifically, we fix the single-document question generator and train a transformer-based document \textit{coordinator} 
% in isolation.
% The RL training learns generator-specific attention to coordinate the current token generation by optimizing a reward based on retrieval statistics from a pre-trained ranker. 

% \subsection{Document-specific GPT-2 Generator}
{\setlength{\parindent}{0cm}
\textbf{Document-specific Generator: }} At the first pre-training stage, we load the publicly available GPT-2 \citep{radford2019language} model as our underlying document-specific generator. 
The GPT-2 model leverages massive out-domain data and serves as a good initialization to generate grammatical and informative question. 
% For configuration details, see \url{https://github.com/openai/gpt-2}.
Then, we further fine-tune the language model on MS-MARCO \citep{DBLP:conf/nips/NguyenRSGTMD16} \textit{selected} document as an input 
% followed by a special separator
and the corresponding question as an output. \newline 
% The generator will be fixed during the RL training.

% At every generation step, the previously generated words are concatenated to all documents as inputs to the generator instances, which yield updated hidden states and output distributions for the next common word. \newline

% \subsection{Ranking-based Rewards Computation}\label{reward}
%by ranking for \st{unsupervised} Learning Coordinator Policy
{\setlength{\parindent}{0cm}
\textbf{Ranking-based Rewards: }\label{reward}}
% \yz{make this paragraph as concise as possible}
Before moving to the RL training, we first describe calculating the reward signal based on retrieval statistics from a BERT-based ranker \citep{DBLP:journals/corr/abs-1901-04085} ($Ranker$), a state-of-the-art model \footnote{http://www.msmarco.org/leaders.aspx} in the MS-MARCO document retrieval task \citep{DBLP:conf/nips/NguyenRSGTMD16}
, is trained to rank (document, question) pairs. 
This ranker assigns high scores for \textit{true positive} document and question pairs. 
% The trained model is publicly available and achieved state-of-the-art results in the MS-MARCO Passage Retrieval task .
% For evaluating the generated questions, we assume the ranker as the oracle since it achieves good performance on the challenging retrieval task.
We assume the ranker delivers an accurate reward signal since it achieves good performance on the challenging MARCO retrieval task, which covers a vast range of general topics. 
Let $\tilde{q}$ be the generated question from the underlying generator block and coordinator with the positive and negative document sets ($\dpos$ and $\dneg$) as the input. 
\begin{align}
     Ranker(d, \tilde{q}) &= \text{score} \in (0,1) \\
     \nonumber &\hspace{0.2in}\forall d \in \dpos,\dneg
\end{align}
We pair $\tilde{q}$ with each of the documents in the positive and negative set, and evaluate the question-document pairs through the ranker for answer-relevancy.
% These retrieval scores for each document that lie in $(0,1)$ are sorted in descending order.
Using the scores and their memberships in $\dpos$ or $\dneg$, we compute retrieval statistics, such as \textit{Precision@10} and \textit{mean-Average-Precision (mAP)} \citep{zhu2004recall} which are candidate non-differentiable rewards $R$. \newline

% \subsection{Inter-document Coordinator}
{\setlength{\parindent}{0cm}
\textbf{Training an Inter-generator Coordinator via RL: }}
Next, we train a coordinator system using policy gradient to optimize the reward described above. 
The separation between the generator and the coordinator aims to ease the RL training by significantly reducing the action space. 
% To the best of the authors' knowledge, this RL procedure has not been employed in question generation before. \yz{@Woonsang, can we say this?}

Note that the generator model is fixed during this stage. We find that using RL to train the entire generating pipeline yields large variance since the action space is large and the auto-regressive nature of the generation process further amplifies such variance. 
Therefore, we fix the underlying generator component and then on top of multiple instances of the underlying generator, we stack our coordinator model, which is trained using RL in isolation.
Instead of training \emph{both} token-level GPT-2 and document-level coordinator over multiple GPT-2 instances using RL, only the coordinator is trained using RL, which structure dramatically reduces variance.

The coordinator is a transformer-based \citep{vaswani2017attention} model to utilize its superior attention capabilities across input documents.% with damped contrastive distribution.
Unlike Transformer Decoder \citep{liu2018generating}, there is no causal mask.
Instead, the coordinator model uses the hidden states updated every decoding time from the underlying fine-tuned GPT-2 \citep{radford2019language} language model generators. 

We add learned \textit{cluster embedding} $c_i$ to the input document hidden states $h_i$, similar to learned positional embedding \citep{devlin2018bert}, to indicate whether the source document $i$ is in $\dpos$ or $\dneg$.
\begin{equation}
x_i^0 = h_i + c_i
\end{equation}
The coordinator model consists of $n$ recurrent transformers blocks \citep{vaswani2017attention}, followed by three different feed-forward layers ($\text{FF}_w,\text{FF}_v$, and $\text{FF}_z$) to output $w$, $v$, and $z$.
\begin{align}
x^k&=\text{Add-Norm}(u,\text{FF}_x(u))\\
u&=\text{Add-Norm}(x^{k-1},\text{MultiHead}(x^{k-1}))\\
&\hspace{0.5cm}\text{for } k=1,\dots,n \nonumber \\
w&=\text{FF}_w(x^n)\\
v&=\text{FF}_v(x^n)\\
z&=\text{FF}_z(x^n)
\end{align}
$w$ and $v$ are the 10-dimensional attention weights that sum to 1.0 among the positive documents $\dpos$, and negative documents $\dneg$.

$z$ parametrizes $\eta$ in how much the coordinator model penalizes, or sometimes reinforces, weighted average of decoding distributions from the negative set $\dneg$. 
$\eta$ is a simple heuristic variation of $\tanh$ such that the image lies in $(-1,0.5)$ for all real numbers $\mathbb{R}$.
Thus, $\eta$ is a damped penalization coefficient.
\begin{equation}
\eta(z) = -\frac{e^{2z}-0.5}{e^{2z}+1} \hspace{.05in} \in (-1,0.5) \hspace{.2in}\forall z \in \mathbb{R}
\end{equation}

Given $w$, $v$, and $z$, we obtain the final question decoding distribution at test time $t$. 
\begin{equation}\label{common_pi}
\pi^{t}_{\theta} = \frac{1}{C}\Big[\sum_{i \in \dpos} w^{t}_{i,\theta} \pi^{t}_i - \eta(z^{t}_\theta) \cdot \sum_{i \in \dneg} v^{t}_{i,\theta} \pi^{t}_i \Big]_{+}
\end{equation}
where $\theta$ is the coordinator's parameters, the subscript $+$ is a ReLU \citep{Nair:2010:RLU:3104322.3104425} operator that selects non-negative weighted tokens, and $C$ is the normalizing factor that converts it into a distribution.
The concatenations of each input document in $\dpos$ and $\dneg$, EOS token, and partially decoded question word sequence are used to obtain new hidden states and next decoding distributions.
The decoding process is repeated until the generation is complete. \newline

% \subsection{Policy Gradient Loss}
{\setlength{\parindent}{0cm}
\textbf{Policy Gradient Loss: }}
The policy gradient loss is defined as follows:
\begin{equation}
\begin{split}
    \mathcal{L}_{\text{PG}}(\theta) &= - \mathbb{E} \Big[ \big( R(\tilde{q} \vert \dpos,\dneg) - R_{\text{baseline}}\big) \\&\cdot \sum_{t} \log \pi^{t}_{\theta}(o_t\vert \tilde{q}_{<t},G, \dpos,\dneg) \Big]
\end{split}
\end{equation}
With a complete generation $\tilde{q}$, a terminal retrieval statistics reward is computed from the $Ranker$ scores and score memberships, noted as $R(\tilde{q} \vert \dpos,\dneg)$.
This reward weights the sum of log-likelihoods of generating the observed words $o_t$ given the generation so far $\tilde{q}_{<t}$, from the underlying generator $G$, and the two document sets. We use \textit{oracle} questions as the policy gradient baseline for variance reduction in $R_{\text{baseline}}$. Results using a different policy gradient baseline are in Appendix.\newline
%Notice that the final decoding distribution (or policy) $\pi_{\theta}$ is a function of the coordinator's parameters $\theta$ and no generator $G$ parameters are modified. \newline

% \subsection{Set-induced Contrastive Regularization
{\setlength{\parindent}{0cm}
\textbf{Set-induced Contrastive Regularization: }}
We further propose an auxiliary to provide richer signals when optimizing the coordinator model. 
The intuition is that we would like to encourage the coordinator model to generate questions \textit{toward} the positive set $\dpos$ relative to the negative set $\dneg$. 
We name the regularizer as \emph{Set-induced} Contrastive Regularization (SCR)
because the decoding distributions from $\dpos$ and $\dneg$ guide the coordinator to learn to make contrasts between the two sets.
Although the decoding distributions from $\dpos$ and $\dneg$ are not gold supervision signals, modifying distributional distance toward or away from them helps regulate specificity to $\dpos$.
The former idea can be formulated as \textit{minimizing} the KL-divergence, evaluated at timestep $t$: 
\begin{dmath}
\min_{\theta} \mathcal{L}_{\text{KL},t}^{\text{pos}}(\theta)=\min_{\theta} \sum_{i\in\dpos} \Big[D_{\text{KL}}(\pi^t_{\theta} \vert\vert  \pi^t_i) + D_{\text{KL}}(\pi^t_i \vert\vert  \pi^t_{\theta}) \Big]
\end{dmath}
We minimize both the forward and the reverse KL divergence since the forward KL does not penalize high mass of $\pi_{\theta}$ where $\pi_i$ does not. 
Likewise for the reverse KL.
On the other hand, the latter idea can be formulated as \textit{maximizing} the KL-divergence against the negative set, evaluated at time step $t$:
\begin{dmath}
\max_{\theta} \mathcal{L}_{\text{KL},t}^{\text{neg}}(\theta)= \max_{\theta} \sum_{i\in\dneg} \Big[D_{\text{KL}}(\pi^t_{\theta} \vert\vert \pi^t_i) + D_{\text{KL}}(\pi^t_i \vert\vert \pi^t_{\theta}) \Big]
\end{dmath}
However, we need to \textit{cap} the negative set penalty rather than na\"ively maximizing it, more restrictively if the positive set and the negative sets are semantically close. 
Intuition is that if the KL divergence against the negative set is too large, then we do not penalize further.
Therefore, we define our contrastive regularization function as follows:
\begin{dmath}
  \mathcal{L}_{\text{SCR}}(\theta)=
  \frac{1}{T} \sum_{t=1}^{T} \Big[ \mathcal{L}_{\text{KL},t}^{\text{pos}}(\theta) - \mathcal{L}_{\text{KL},t}^{\text{neg}}(\theta)\cdot \mathds{1}_{\nu_t \cdot \mathcal{L}_{\text{KL},t}^{\text{neg}}(\theta) < \mathcal{L}_{\text{KL},t}^{\text{pos}}(\theta)} \Big]
\end{dmath}
where $T$ is the length of the completed generation, and $\nu_t$ is the similarity measure between positive and negative sets at decoding time $t$.
Specifically,
\begin{dmath}
\nu_t=\cos\text{sim}\bigg(\frac{1}{\vert \dpos \vert}\sum_{i\in\dpos} \pi^t_i,\frac{1}{\vert\dneg \vert}\sum_{i\in\dneg} \pi^t_i\bigg)
\end{dmath}

\begin{table*}[t!]
% \caption{Retrieval performance. 
% ``Out-Sample IR'' refers to the evaluation data sample that consists of 10+10 documents $\dpos$ and $\dneg$. 
% ``Search-Engine Augmented IR'' refers to augmenting the out-sample into 100 documents in total through Lucene.
% }
%The results for the RNN-based MSQG are directly from \citet{cho_qgen_2019}.}
\resizebox{\textwidth}{!}{%
\begin{tabular}{l|cccc|ccccc}
 & \multicolumn{4}{c|}{\textbf{Out-Sample IR}} & \multicolumn{5}{c}{\textbf{Search-Engine Augmented IR}} \\ \hline
\textbf{Model} & mAP & RPrec & MRR (=MRR@10) & nDCG & mAP & RPrec & MRR & MRR@10 & nDCG \\ \hline
Top-TFIDF @100 & 0.416 & 0.533 & 0.696 & 0.545 & 0.113 & 0.0588 & 0.0260 & 0.0050 & 0.181 \\
Top-Frequent @100 & 0.680 & 0.742 & 0.921 & 0.779 & 0.171 & 0.129 & 0.0404 & 0.0119 & 0.204 \\ \hline
MSQG (Cho et al. '19) & - & - & - & - & - & - & 0.0704 & \textbf{0.0441} & 0.234 \\ \hline
$\text{MSQG}_{GPT2}$ & 0.713 & 0.763 & 0.945 & 0.804 & 0.245 & 0.217 & 0.0714 & 0.0400 & 0.240 \\
$\text{MSCQG}_{SCR}$ & 0.751 & 0.790 & 0.974 & 0.836 & 0.258 & 0.234 & 0.0745 & 0.0420 & 0.245 \\
$\text{MSCQG}_{PG}$ & 0.753 & 0.791 & 0.978 & 0.838 & 0.256 & 0.232 & 0.0742 & 0.0421 & 0.244 \\
$\text{MSCQG}_{PG+SCR}$ & \textbf{0.767} & \textbf{0.803} & \textbf{0.981} & \textbf{0.849} & \textbf{0.265} & \textbf{0.242} & 0.0748 & 0.0420 & 0.245 \\
$\text{MSCQG}_{PG+SCR+H}$ & 0.765 & 0.800 & 0.976 & 0.847 & 0.262 & 0.239 & \textbf{0.0759} & 0.0434 & \textbf{0.246} \\ \hline
\emph{Oracle} Questions for $\dpos$& 0.759 & 0.797 & 0.976 & 0.842 & 0.292 & 0.273 & 0.0846 & 0.0495 & 0.256 \\ 
% \hline
\end{tabular}%
}
\caption{Retrieval performance. 
``Out-Sample IR'' refers to the evaluation data sample that consists of 10+10 documents $\dpos$ and $\dneg$. 
``Search-Engine Augmented IR'' refers to augmenting the out-sample into 100 documents in total through Lucene.
}
\label{overall_retrieval}
\end{table*}

\begin{table*}[t!]
% \caption{Comparison against the \emph{oracle} MARCO questions for $\dpos$. Since retrieval scores cannot give a complete picture of the generation, we aim to understand how close the generations are in terms of various metrics. The numbers show that our proposed model generates questions similar to the \emph{oracle} MARCO questions. \textbf{Notations:} BL for BLEU; ST for Skip-Thought similarity; EM for Embedding Mean similarity; VE for Vector Extrema similarity; and GM for Greedy Matching. }
\resizebox{\textwidth}{!}{%
\begin{tabular}{l|cccc|cc|ccccc}
 & BL-1 & BL-2 & BL-3 & BL-4 & METEOR & ROUGE\_L & CIDEr & ST & EM & VE & GM \\ \hline
\emph{Oracle} Question for $\dneg$ & 0.449 & 0.291 & 0.177 & 0.100 & 0.215 & 0.428 & 1.076 & 0.547 & 0.766 & 0.617 & 0.697 \\
Top-TFIDF @100 & 0.253 & 0.157 & 0.104 & 0.075 & 0.195 & 0.339 & 1.174 & 0.470 & 0.747 & 0.575 & 0.671 \\
Top-Frequent @100 & 0.438 & 0.328 & 0.260 & 0.217 & 0.281 & 0.476 & 2.684 & 0.573 & 0.799 & 0.682 & 0.735 \\
$\text{MSQG}_{GPT2}$ & 0.457 & 0.313 & 0.207 & 0.139 & 0.282 & 0.494 & 1.993 & 0.563 & 0.814 & 0.705 & 0.768 \\
$\text{MSCQG}_{SCR}$ & 0.501 & 0.363 & 0.260 & 0.193 & 0.303 & 0.535 & 2.533 & 0.604 & 0.829 & 0.729 & 0.786 \\
$\text{MSCQG}_{PG}$ & 0.562 & 0.418 & 0.310 & 0.234 & 0.304 & 0.565 & 2.702 & 0.630 & 0.844 & 0.734 & 0.798 \\
$\text{MSCQG}_{PG+SCR}$ & \textbf{0.589} & \textbf{0.449} & \textbf{0.339} & \textbf{0.262} & \textbf{0.323} & \textbf{0.591} & \textbf{2.994} & \textbf{0.647} & \textbf{0.858} & \textbf{0.759} & \textbf{0.815} \\
$\text{MSCQG}_{PG+SCR+H}$ & 0.573 & 0.436 & 0.330 & 0.255 & 0.321 & 0.583 & 2.946 & 0.641 & 0.851 & 0.752 & 0.808
\end{tabular}%
}
\caption{Comparison against the \emph{oracle} MARCO questions for $\dpos$. Since retrieval scores cannot give a complete picture of the generation, we aim to understand how close the generations are in terms of various metrics. The numbers show that our proposed model generates questions similar to the \emph{oracle} MARCO questions. \textbf{Notations:} BL for BLEU; ST for Skip-Thought similarity; EM for Embedding Mean similarity; VE for Vector Extrema similarity; and GM for Greedy Matching. }
\label{pseudo_closeness}
\end{table*}

% \subsection{Entropy Loss}%$\bm{\left(H\right)}$}
{\setlength{\parindent}{0cm}
\textbf{Negative Entropy Loss: }}
We add negative entropy loss $\mathcal{L}_{\text{H}}$ across the attention weights $w$ and $v$, averaged over $T$ to encourage the model attend to all the documents rather than attend to a small subset of the documents and risk losing positive and negative set representational information.
\begin{dmath}
    \mathcal{L}_{\text{H}}(\theta) = \frac{1}{T} \sum_{t=1}^{T} \Big[
    \sum_{i \in \dpos} w^t_{i,\theta}\log w^t_{i,\theta} +
    \sum_{i \in \dneg} v^t_{i,\theta}\log v^t_{i,\theta}
    \Big]
\end{dmath}
{\setlength{\parindent}{0cm}
We finally optimize for the following loss:
\begin{dmath}
\mathcal{L}(\theta)=
\lambda_{1} \mathcal{L}_{\text{PG}}(\theta)+
\lambda_{2} \mathcal{L}_{\text{SCR}}(\theta)+
\lambda_{3} \mathcal{L}_{\text{H}}(\theta)
\end{dmath}
}
where $\lambda_{1,2,3}$ are the scaling hyper-parameters.

\section{Experiments}\label{experiments}
\textbf{Dataset: } We use the MS-MARCO Q\&A dataset \citep{DBLP:conf/nips/NguyenRSGTMD16} where for the Bing query $q$, we consider the top-10 retrieved documents as our positive set $\dpos$. To get our negative set $\dneg$, we use the Conversational Search\footnote{https://github.com/microsoft/MSMARCO-Conversational-Search} dataset, which contains additional annotations for the same MS-MARCO Bing queries, to find a query $q'$ that is similar to $q$ yet not a paraphrase and consider the top-10 documents retrieved for $q'$ as our negative set $\dneg$.
In total, we gather 100K train/10K dev/10K eval data points.
Details of the pre-processing, usage of additional annotations from the secondary dataset, and experimental configuration are in Appendix.\newline

\begin{figure*}[t!]%
    \centering
    \begin{minipage}[b]{.5\textwidth}
    {\includegraphics[width=1.0\linewidth]{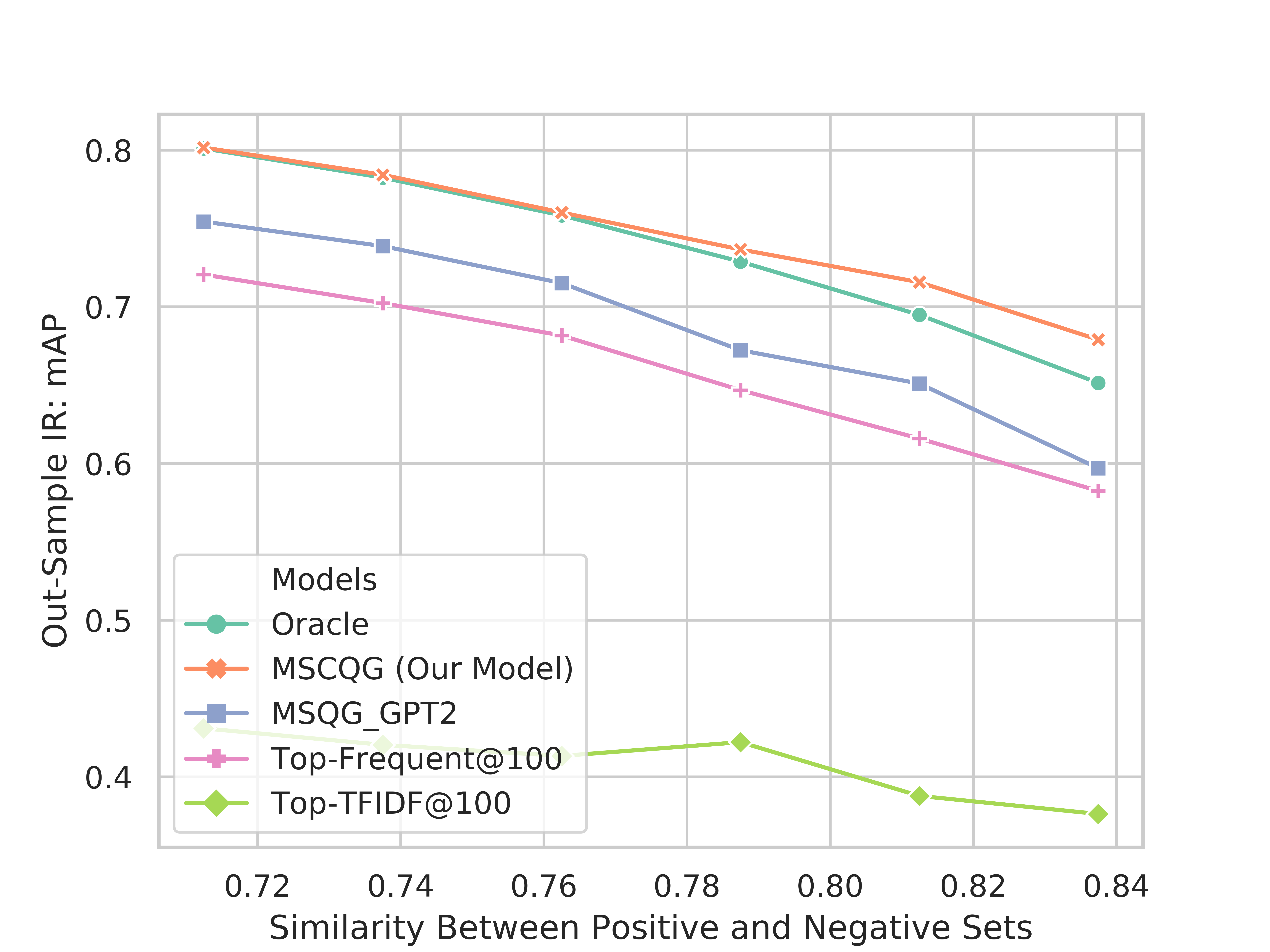}}
    \caption{Out-Sample IR: mAP among $\dpos$ and $\dneg$}\label{outsample_ir}
    \end{minipage}\hfill%\qquad
    \begin{minipage}[b]{.5\textwidth}
    {\includegraphics[width=1.0\linewidth]{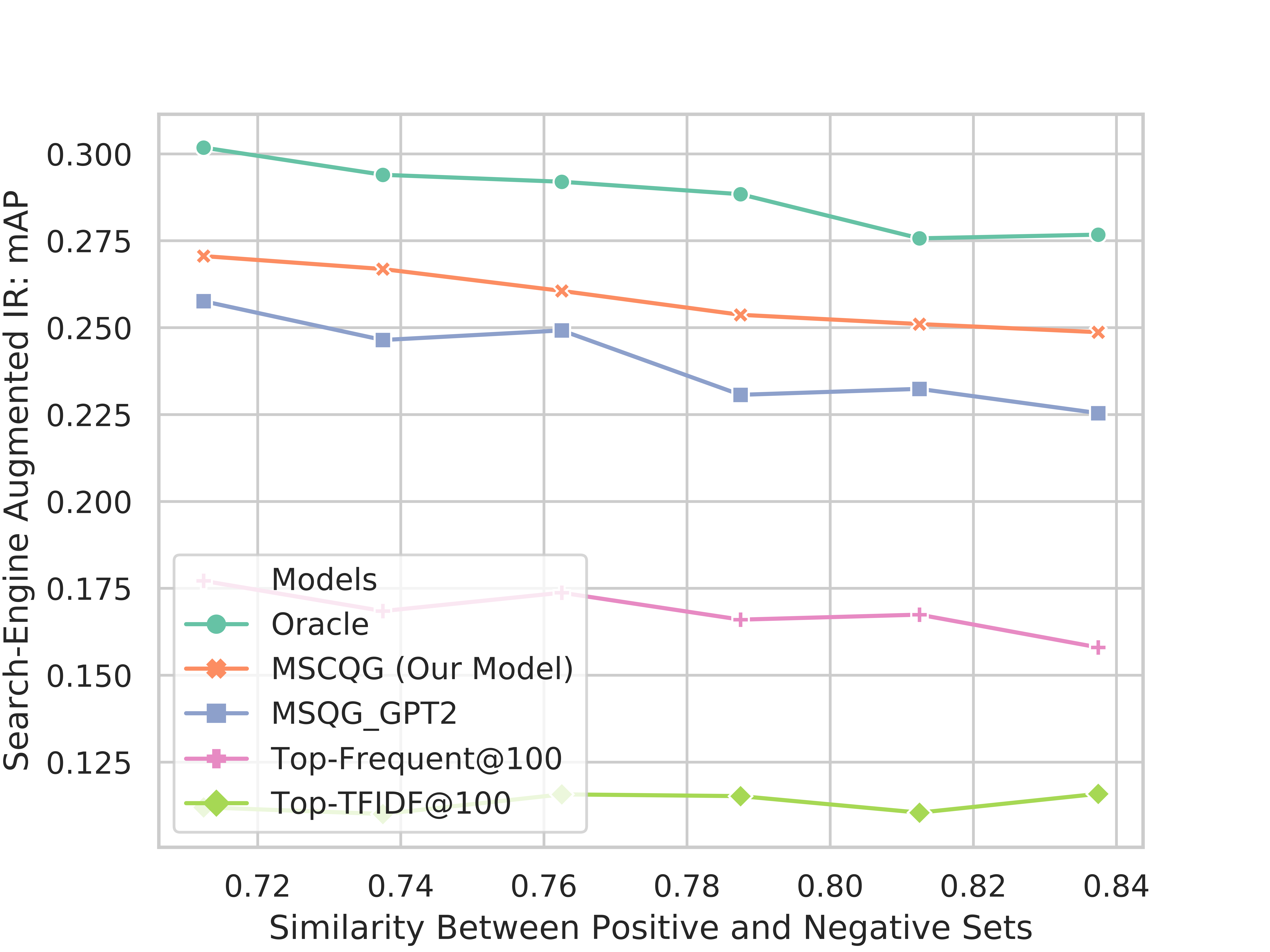}}
    \caption{Search-Engine Augmented IR: mAP}\label{lucene_ir}
    \end{minipage}
    \caption*{Figure~\ref{outsample_ir} shows that our model MSCQG$_{PG+SCR+H}$ outperforms the \emph{oracle} questions by a small margin on the Out-Sample IR. In the larger retrieval evaluation using Lucene, it performs subpar against the \emph{oracle} questions, but performs significantly better than all the considered baseline models, shown in Figure~\ref{lucene_ir}.}%
\end{figure*}

{\setlength{\parindent}{0cm}
\textbf{Automatic evaluation: }}
The generated questions are evaluated through standard retrieval-based metrics: MRR and MRR\@10 \citep{voorhees1999proceedings, radev-etal-2002-evaluating}, nDCG \citep{Jarvelin:2002:CGE:582415.582418}, precision, mAP.
These metrics are computed from the 10 positive and 10 negative document sets (=: \textit{Out-Sample IR}).
In addition, as a standardized evaluation routine in the MS-MARCO Retrieval task, for each generated question, we use Lucene\footnote{https://lucene.apache.org/} to retrieve the most relevant 100 MARCO documents via BM25 \citep{Robertson:2009:PRF:1704809.1704810}, and use the retrieved document set and a trained model to rank (document, generated question) pairs, thus compute the retrieval statistics (=:\textit{ Search-Engine Augmented IR}).

The generated questions are also evaluated in terms of BLEU \citep{papineni2002bleu}, ROUGE \citep{Lin:2003:AES:1073445.1073465}, METEOR \citep{banerjee2005meteor}, CIDEr \citep{vedantam2015cider}, Greedy Matching \citep{rus2012comparison}, Skip-Thought \citep{kiros2015skip}, Embedding Average \citep{kenter2016siamese} and Vector Extrema \citep{forgues2014bootstrapping} cosine similarities.\newline

\begin{table}[]
% \caption{Pairwise comparison and $d^+/d^-$ comparison of human evaluation. M=MSCQG, B=MSQG$_{GPT2}$, O=\emph{Oracle}. Preferences are expressed in percentage (\%). Comparison results are statistically significant ($p<0.01$) unless indicated *.
% Ans., Rel., Flu. and Ovr. denotes \textbf{Ans}werability, \textbf{Rel}evancy and \textbf{Flu}ency and \textbf{Ov}e\textbf{r}all, respectively.
% }
\resizebox{\columnwidth}{!}{%
\begin{tabular}{p{0.5in}ccccccccc}
\hline
 & \multicolumn{3}{c}{Pair (M vs. B)} & \multicolumn{3}{c}{Pair (M vs. O)} & \multicolumn{3}{c}{Pair (B vs. O)} \\ \cline{2-10} 
Criteria & M & B & \multicolumn{1}{c|}{=} & M & O & \multicolumn{1}{c|}{=} & B & O & \multicolumn{1}{c}{=} \\ \hline
Ans. & \textbf{52.2} & 17.5 & \multicolumn{1}{c|}{30.3} & 39.2 & 19.8 & \multicolumn{1}{c|}{\textbf{41.0}} & 32.3 & \textbf{42.5} & \multicolumn{1}{c}{25.2} \\
Rel. & \textbf{53.3} & 18.7 & \multicolumn{1}{c|}{28.0} & 35.2 & 22.2 & \multicolumn{1}{c|}{\textbf{42.7}} & 31.7 & \textbf{44.2} & \multicolumn{1}{c}{24.2} \\
Flu. & \textbf{49.3} & 22.3 & \multicolumn{1}{c|}{28.3} & \textbf{50.8} & 24.7 & \multicolumn{1}{c|}{24.5} & \textbf{43.7} & 32.7 & \multicolumn{1}{c}{23.7} \\ \hline
Ovr. & \textbf{57.5} & 21.3 & \multicolumn{1}{c|}{21.2} & \textbf{49.5} & 27.0 & \multicolumn{1}{c|}{23.5} & 38.3 & \textbf{42.8*} & \multicolumn{1}{c}{18.8} \\ 
% \hline
 & \multicolumn{1}{l}{} & \multicolumn{1}{l}{} & \multicolumn{1}{l}{} & \multicolumn{1}{l}{} & \multicolumn{1}{l}{} & \multicolumn{1}{l}{} & \multicolumn{1}{l}{} & \multicolumn{1}{l}{} & \multicolumn{1}{l}{} \\ \hline
 & \multicolumn{3}{c}{M} & \multicolumn{3}{c}{B} & \multicolumn{3}{c}{O} \\ \cline{2-10} 
Criteria & $d+$ & $d-$ & \multicolumn{1}{c|}{=} & $d+$ & $d-$ & \multicolumn{1}{c|}{=} & $d+$ & $d-$ & \multicolumn{1}{c}{=} \\ \hline
Ans. & \textbf{70.7} & 10.2 & \multicolumn{1}{c|}{19.2} & \textbf{61.2} & 14.0 & \multicolumn{1}{c|}{24.8} & \textbf{67.2} & 13.7 & \multicolumn{1}{c}{19.2} \\ 
Rel. & \textbf{72.2} & 11.3 & \multicolumn{1}{c|}{16.5} & \textbf{62.2} & 16.3 & \multicolumn{1}{c|}{21.5} & \textbf{70.3} & 14.3 & \multicolumn{1}{c}{15.3} \\ \hline
Ovr. & \textbf{72.0} & 10.5 & \multicolumn{1}{c|}{17.5} & \textbf{63.2} & 14.2 & \multicolumn{1}{c|}{22.7} & \textbf{69.0} & 14.0 & \multicolumn{1}{c}{17.0} \\ 
% \hline\hline
\end{tabular}%
}
\caption{Pairwise comparison and $d^+/d^-$ comparison of human evaluation. M=MSCQG, B=MSQG$_{GPT2}$, O=\emph{Oracle}. Preferences are expressed in percentage (\%). Comparison results are statistically significant ($p<0.01$) unless indicated *.
Ans., Rel., Flu. and Ovr. denotes \textbf{Ans}werability, \textbf{Rel}evancy and \textbf{Flu}ency and \textbf{Ov}e\textbf{r}all, respectively.
}
\label{human-eval-exp1}
\vspace*{-3mm}
\end{table}
{\setlength{\parindent}{0cm}
\textbf{Human evaluation: }}
We conduct human evaluation through Amazon Mechanical Turk where we evaluate questions generated by MSCQG, $\text{MSQG}_{GPT2}$, and the \emph{oracle} question in four criteria: \textit{fluency}, \textit{relevancy}, \textit{answerability}, and \textit{overall}. 
First, we randomly select 600 $(d, q_A, q_B)$ tuples where the $d$ is any from $\dpos$ and $q_A, q_B$ from the three questions, and collect responses on which question is preferred over the other.
Secondly, we evaluate 600 $(d^+,d^-, q)$ tuples where given a question, $d^+,d^-$ are randomly chosen from $\dpos$ and $\dneg$. 
This can determine questions' specificity to $\dpos$ relative to $\dneg$.
Each sample is judged by 3 crowd-sourced workers who passed a rigorous spam-detection screening, totalling 3,600 samples to obtain reliable results. For details, see Appendix.
% \vspace*{-5mm}
\subsection{Baseline models}\label{baseline}
{\setlength{\parindent}{0cm}
\textbf{Multi-Source Question Generator: } This model MSQG$_{GPT2}$ is similar to MSQG in \citet{cho_qgen_2019}.
It processes individual documents in parallel through the fine-tuned GPT-2 generator, rather than RNN-based Seq2Seq model in MSQG, and averages the decoding distributions at test time $t$.
\begin{equation}
\pi_{\text{MSQG}_{GPT2}}^t = \frac{1}{\vert pos \vert}\sum_i^{pos} \pi^t_i
\end{equation}
Unlike MSQG in \citet{cho_qgen_2019}, no further heuristic modifications are made to the model.
}\newline

{\setlength{\parindent}{0cm}
\textbf{Top-TFIDF@K: } 
\emph{Why do we not simply retrieve the top question implied by the 10 positive documents?} 
To this end, we design a retrieval baseline using the learned TF-IDF \citep{luhn1957statistical,Jones72astatistical,salton1983introduction} weights.
This baseline re-evaluates the collection of retrieved questions from the corpus, gathered against each document in $\dpos$ using TF-IDF, and retrieves the most relevant question.
For design details, see Appendix.
%For pseudo-code details, see Appendix.
}\newline

{\setlength{\parindent}{0cm}
\textbf{Top-Frequent@K: } Another retrieval model is to find an intersecting subset among all the 10 top-$k$ question sets.
For pseudo-code details, see Appendix. 
}

\begin{figure*}[t!]%
    \centering
    \begin{minipage}[b]{.5\textwidth}
    {\includegraphics[width=1.0\linewidth]{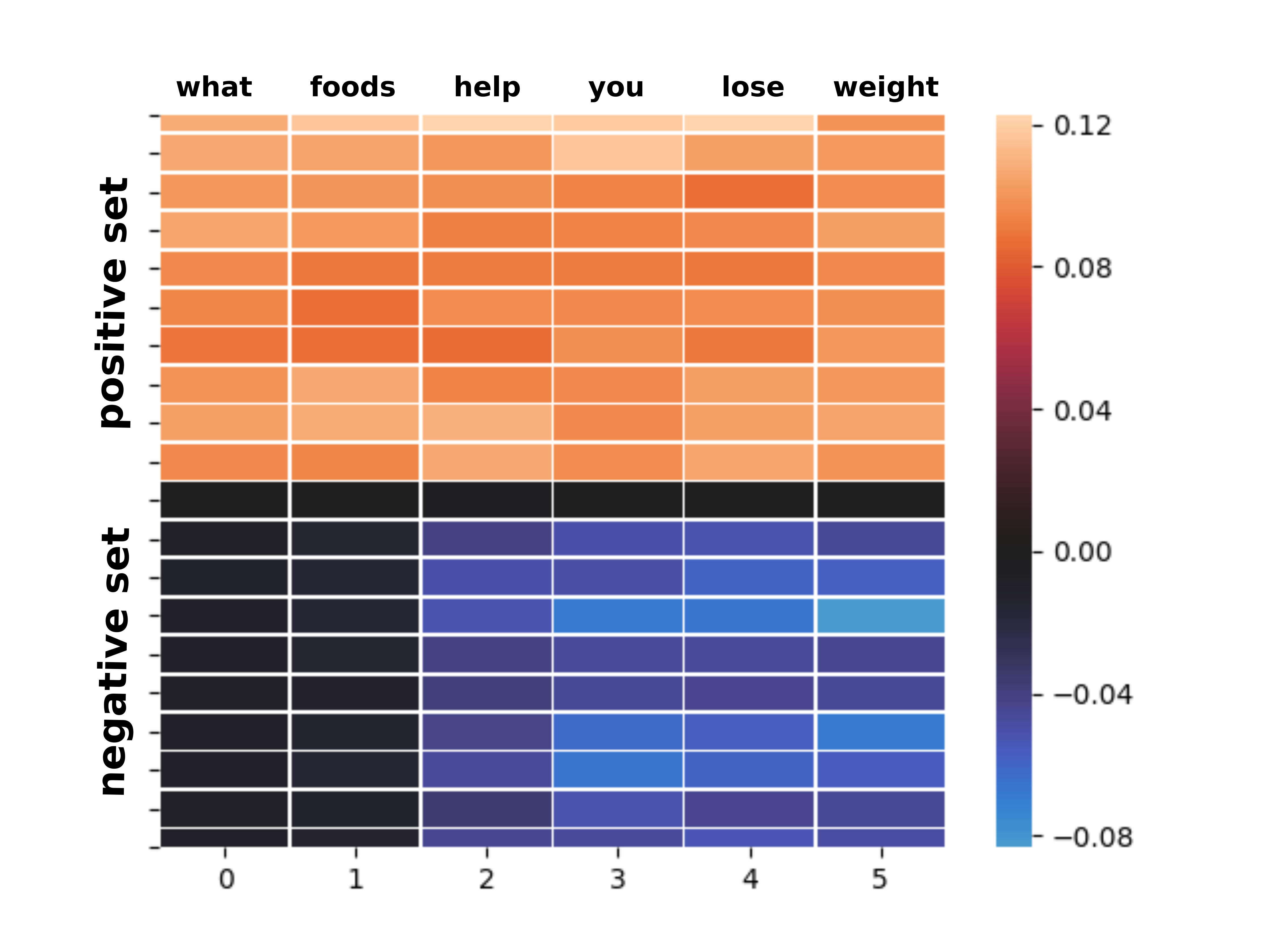}}
    \caption{Example attention weights \#1}\label{gradual}
    \end{minipage}\hfill%\qquad
    \begin{minipage}[b]{.5\textwidth}
    {\includegraphics[width=1.0\linewidth]{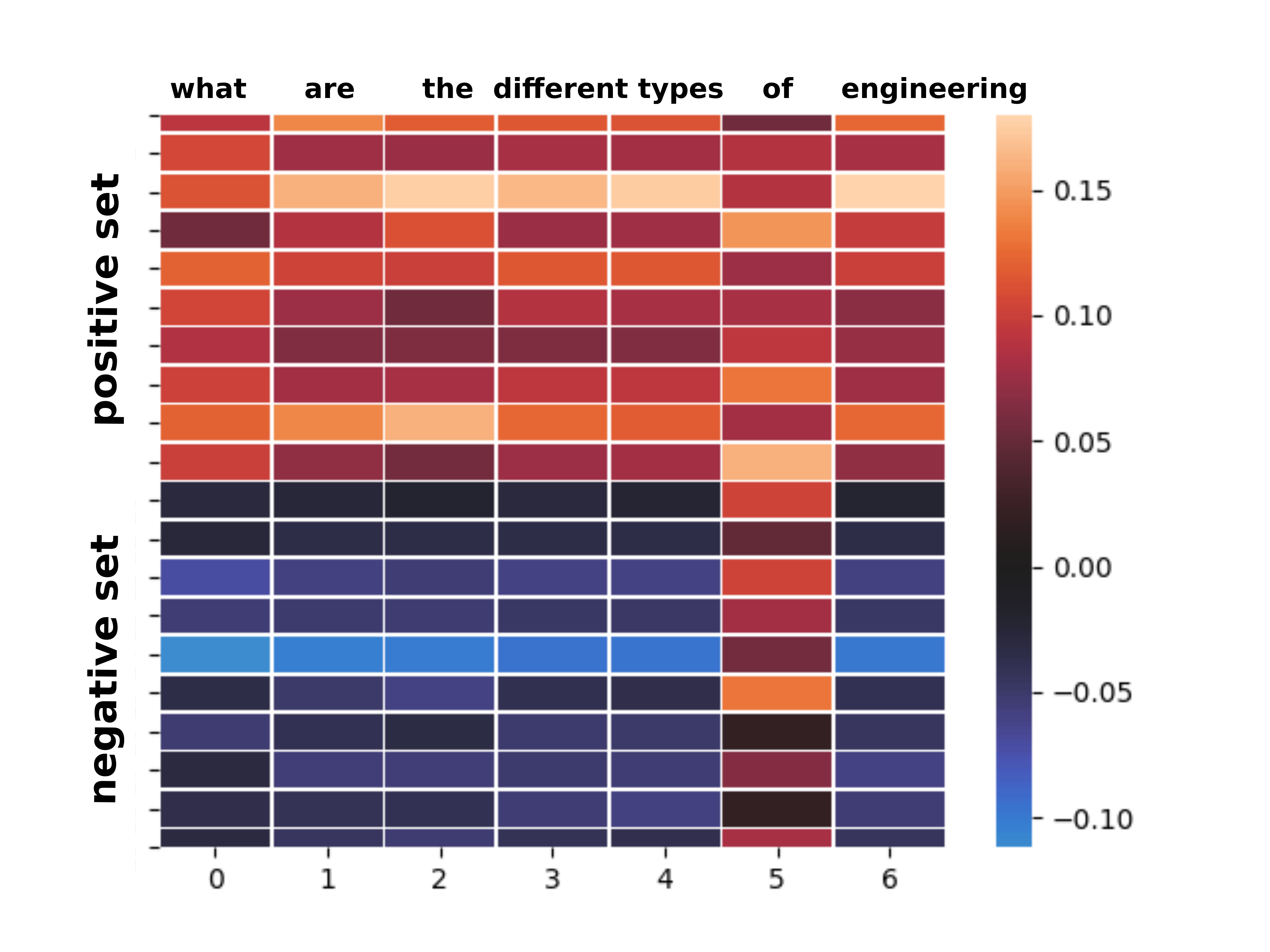}}
    \caption{Example attention weights \#2}\label{selective}
    \end{minipage}
\caption*{Visualization of sequential attention weights. 
In the vertical axis, 0-9 indices indicate documents in $\dpos$, and 10-19 in $\dneg$. 
For explanation on $\dneg$ weights $v$, see analysis below. 
Figure~\ref{gradual} shows that the model learns to push the sequential generation semantics more toward $\dpos$ by gradually penalizing $\dneg$.
Figure~\ref{selective} shows that frequent and semantically less distinguishing words such as \emph{`of'} are encouraged even by $\dneg$, which empirically aligns with our intuition for TF-IDF. 
}%%
\end{figure*}

\subsection{Results and Analysis}
% \newline

{\setlength{\parindent}{0cm}
\textbf{Model comparison and ablation study:}}
For simplicity, we abuse the term \emph{oracle} by calling the ground-truth question that retrieves $\dpos$ when constructing the dataset as the \emph{oracle} question.
However, these questions are not gold questions as they might not be the most relevant and specific questions to the given positive and negative sets. 

Table~\ref{overall_retrieval} shows that our proposed model is effective at generating questions given multiple documents. 
In particular, it shows that policy gradient or set-induced contrastive regularization alone is effective in improving performance. 
The coordinator performs better when optimized for both policy gradient and regularization objectives.

The retrieval results for the questions that initially clustered $\dpos$ sets are presented. 
Note that these are not \emph{gold} questions because in most cases not all the retrieved documents in $\dpos$ answer the questions.
For clarity of our presentation, we abuse the term and name them as \emph{oracle} questions.
Search-engine augmented IR evaluation shows that our methods are upper-bounded by the \emph{oracle} MARCO questions.

Entropy regularization improves the search-engine augmented IR scores, in particular, MRR. 
However, it is not crucial as supplemented by Table~\ref{pseudo_closeness}. 
For additional results, see Table~\ref{additional_overall_retrieval} in Appendix. \newline

{\setlength{\parindent}{0cm}
\textbf{Model performance v.s. similarities between $\dpos$ and $\dneg$:}}
$\cossim(\dpos,\dneg)$ is approximated using the \emph{oracle} questions that are available in the dataset.
The similarity is computed by the cosine similarity of the two GEN-Encoder \citep{Zhang:2019:GIR:3331184.3331198} representations. 
Figures~\ref{outsample_ir} and~\ref{lucene_ir} show that our model generated questions are more grounded on $\dpos$ than the baseline model generations. 
The more similar the two sets $\dpos$ and $\dneg$, the more difficult for the models, even humans, to distinguish which document is more relevant, if not answerable, given the generated question. 
The model outperforms the baseline model uniformly across different similarities between $\dpos$ and $\dneg$. \newline

{\setlength{\parindent}{0cm}
\textbf{Role of $\dneg$ by visualizing $w$, $v$, and $z$: }}
Figures~\ref{gradual} and~\ref{selective} show that our model MSCQG learns to gradually penalize $\dneg$ as it sequentially generates words that are more grounded on $\dpos$. 
Notice the roughly uniform weights across $\dpos$ but increasing penalization weights across $\dneg$, in decoding time.

$\eta$, which is controlled by the $z$, is learned to encourage, rather than discourage, certain words during decoding.
The displayed $\dneg$ weights $v$ are multiplied by the dampening factor $-\eta(z)$ for interpretation purposes, thus it does not necessarily sum to 1, see equation~\ref{common_pi}.
We observe that words that are not semantically distinguishing between $\dpos$ and $\dneg$, are encouraged by the coordinator to maintain readability.
For example, the weights of the word \emph{of} is mostly non-negative, whereas weights for other words are mostly negative.
This indicates that the coordinator learns to selectively activate/suppress decoding of certain words by coordinating information from $\dpos$ and $\dneg$. \newline

{\setlength{\parindent}{0cm}
\textbf{Human judgments:}} Table~\mbox{\ref{human-eval-exp1}} shows that our model significantly outperforms the strong baseline in every aspect. 
%The results are statistically significant as marked, drawn from the large number of evaluations. 
Furthermore, we draw a more favorable conclusion toward our model-generated questions when compared against the \emph{oracle} questions than from the automatic metrics, which are approximate yet reasonable metrics. The pairwise agreement between judges is 54\% $\pm$ 1\%. The Cohen's Kappa score is 0.19 $\pm$ 0.01. Note that this is a reasonable number given the \textit{``same''} or ambiguous option in pairwise comparisons. $Ranker$ achieves a relatively high Pearson correlation of 0.6 with respect to human evaluation. For details, see Appendix.

\section{Related Work}\label{related}   

{\setlength{\parindent}{0cm}
\textbf{Multi-Source Encoder-Decoder: } 
Ensemble set induction mechanism \citep{Rokach2010} has been widely  applied to neural machine translation (NMT) tasks \citep{bojar2014findings}.
\citet{DBLP:journals/corr/FiratSAYC16} introduced a new type of ensemble of NMT systems which take inputs as multiple sentences in different languages and output a translation into a single language.
Each NMT system is trained on a mono-lingual source to target language translation dataset.
\citet{garmash-monz-2016-ensemble} further developed the multi-source encoder-decoder framework for multi-lingual NMT systems, by learning to assign uneven attention weights, called \emph{expert combination weights}. 
To handle multi-source input, we take a similar multi-source encoder-decoder approach for our coordinator model.
For such multi-lingual translation tasks, the target translation is available. 
However, in our task of generating multi-document questions, the target \emph{does not exist} which makes it more challenging, thus we train via RL, rather than supervised learning.
}\newline

{\setlength{\parindent}{0cm}
\textbf{Question Generation: } 
Most prior work on question generation has been on single document i.e. given a document and an answer phrase in the document, generate a question that is answered by the answer phrase \citep{heilman2011automatic, rus2010first}. 
For a survey, see \citet{DBLP:journals/corr/abs-1905-08949}.
However, in our work, we aim to generate a multi-document question that is answerable by multiple input documents.
%, similar to \citet{cho_qgen_2019}, whom introduced common question generation from multi-documents.
% Recently, sequence-to-sequence based neural network models have defined the state-of-the-art for question generation \citep{du2017learning, duan2017question}. 
% Our generator model, on the other hand, is based on the more recent GPT-2 \citep{radford2019language} generation model, and this forms the underlying component of our question generating system.
\citet{fan-etal-2018-reinforcement} propose a visual question generation model to generate natural questions about images using reinforcement learning where they use naturalness and human-like as reward signals. 
In our work, we use retrieval statistics, similar to \citet{nogueira-cho-2017-task}, derived from a document-question ranker as the reward for training our coordinator model in isolation, rather than the entire generating pipeline. 
}\newline

{\setlength{\parindent}{0cm}
\textbf{Contrastive learning in NLP: } 
Contrastive learning has been widely used in NLP \citep{smith-eisner-2005-contrastive,JMLR:v12:collobert11a,NIPS2013_5071,hjelm2019learning,Deng2020Residual}. Broadly, contrastive learning methods differentiate observed data from artificial negative examples. 
\citet{pmlr-v9-gutmann10a} leverages the Noise Contrastive Estimation (NCE) metric to differentiate the target sample from noise samples. Negative Sampling proposed by \citet{mikolov2013distributed} is a simplified variation of NCE loss. Recently, contrastive learning has also been employed in learning sentence representations \citep{clark2020electra}. 
%Our contrastive learning approach is fundamentally different from above, which learn representations. 
To our best knowledge, we are the first to leverage contrastive learning and establish set-induced penalization in the context of question generation.
}

% \section{Preliminary}\label{prelim}
% \citet{cho_qgen_2019} introduced common question generation from multi-documents. 
% The authors employ an information aggregation mechanism by linearly interpolating the output distributions of each decoder. 
% A recurrent neural network \citep{rumelhart1988learning,58337} sequence-to-sequence model \citep{seq2seq,DBLP:journals/corr/BahdanauCB14} is trained from one document (input) and one question (output) that is \textit{answerable} by the input.
% Using multiple instances of the model, individual inferences are made from $N$ documents after which decoding distributions $\pi_{i,t}$ ($i\in\{1,\cdots,N\}$) are averaged to generate a common question word at time $t$, followed by additional heuristics.
% Overall, the generated questions are relevant to the documents, but relatively generic. This is because the training objective only requires the generation to be answerable by all documents, thus a generic question that covers broad topics is not penalized. 

\section{Conclusion}
We proposed a novel coordinator model that can generate questions that are more grounded on documents of interest.
This coordinator model consists of transformer blocks, and is trained through reinforcement learning and an effective auxiliary: Set-induced Contrastive Regularization.

The rewards are derived from a publicly available state-of-the-art pre-trained ranker (Section~\ref{model}) to compute retrieval statistics among $\dpos$ and $\dneg$. 
Our novel contrastive regularization induces generations to be more specific to $\dpos$ than to $\dneg$ while limiting the effect of $\dneg$ in a principled manner by accounting for their semantic similarity.

We evaluate a generated question from each model by assessing how many of the input $\dpos$ documents among a pool of relevant documents it can retrieve, based on the (document, question) ranker that is trained on the same wide-ranging domain.
For a comprehensive automatic evaluation of the models, retrieval statistics are computed from a larger pool of relevant documents gathered via BM25.  
Experiment results show that our model significantly outperforms previous neural generation as well as strong retrieval baselines in both automatic and human metrics.

Given the promising comprehensive results of the proposed models and training approach, we can extend the framework with appropriate modifications and train via imitation learning algorithms, and this is left for future work.

\section*{Acknowledgements}
We would like to thank Paul Bennett, Xiang Gao, Matthew Richardson, and Michel Galley at Microsoft Research for thoughtful feedback. We would also like to thank the anonymous reviewers for their time and providing helpful suggestions. 

%we plan to dual-submit our work to conferences that allow dual submission.

\nocite{*}
\bibliographystyle{acl_natbib}
\bibliography{eacl2021}

\begin{thebibliography}{120}
\expandafter\ifx\csname natexlab\endcsname\relax\def\natexlab#1{#1}\fi

\bibitem[{Aliannejadi et~al.(2019)Aliannejadi, Zamani, Crestani, and
  Croft}]{10.1145/3331184.3331265}
Mohammad Aliannejadi, Hamed Zamani, Fabio Crestani, and W.~Bruce Croft. 2019.
\newblock \href {https://doi.org/10.1145/3331184.3331265} {Asking clarifying
  questions in open-domain information-seeking conversations}.
\newblock In \emph{Proceedings of the 42nd International ACM SIGIR Conference
  on Research and Development in Information Retrieval}, SIGIR’19, page
  475–484, New York, NY, USA. Association for Computing Machinery.

\bibitem[{Bahdanau et~al.(2014)Bahdanau, Cho, and
  Bengio}]{DBLP:journals/corr/BahdanauCB14}
Dzmitry Bahdanau, Kyunghyun Cho, and Yoshua Bengio. 2014.
\newblock \href {http://arxiv.org/abs/1409.0473} {Neural machine translation by
  jointly learning to align and translate}.
\newblock \emph{CoRR}, abs/1409.0473.

\bibitem[{Banerjee and Lavie(2005)}]{banerjee2005meteor}
Satanjeev Banerjee and Alon Lavie. 2005.
\newblock Meteor: An automatic metric for mt evaluation with improved
  correlation with human judgments.
\newblock In \emph{Proceedings of the acl workshop on intrinsic and extrinsic
  evaluation measures for machine translation and/or summarization}, pages
  65--72.

\bibitem[{Bernhard(2010)}]{bernhard-2010-query}
Delphine Bernhard. 2010.
\newblock \href {https://www.aclweb.org/anthology/C10-2007} {Query expansion
  based on pseudo relevance feedback from definition clusters}.
\newblock In \emph{Coling 2010: Posters}, pages 54--62, Beijing, China. Coling
  2010 Organizing Committee.

\bibitem[{Bojar et~al.(2014)Bojar, Buck, Federmann, Haddow, Koehn, Leveling,
  Monz, Pecina, Post, Saint-Amand et~al.}]{bojar2014findings}
Ondrej Bojar, Christian Buck, Christian Federmann, Barry Haddow, Philipp Koehn,
  Johannes Leveling, Christof Monz, Pavel Pecina, Matt Post, Herve Saint-Amand,
  et~al. 2014.
\newblock Findings of the 2014 workshop on statistical machine translation.
\newblock In \emph{Proceedings of the ninth workshop on statistical machine
  translation}, pages 12--58.

\bibitem[{Bordes et~al.(2013)Bordes, Usunier, Garcia-Duran, Weston, and
  Yakhnenko}]{NIPS2013_5071}
Antoine Bordes, Nicolas Usunier, Alberto Garcia-Duran, Jason Weston, and Oksana
  Yakhnenko. 2013.
\newblock \href
  {http://papers.nips.cc/paper/5071-translating-embeddings-for-modeling-multi-relational-data.pdf}
  {Translating embeddings for modeling multi-relational data}.
\newblock In C.~J.~C. Burges, L.~Bottou, M.~Welling, Z.~Ghahramani, and K.~Q.
  Weinberger, editors, \emph{Advances in Neural Information Processing Systems
  26}, pages 2787--2795. Curran Associates, Inc.

\bibitem[{Braslavski et~al.(2017)Braslavski, Savenkov, Agichtein, and
  Dubatovka}]{10.1145/3020165.3022149}
Pavel Braslavski, Denis Savenkov, Eugene Agichtein, and Alina Dubatovka. 2017.
\newblock \href {https://doi.org/10.1145/3020165.3022149} {What do you mean
  exactly? analyzing clarification questions in cqa}.
\newblock In \emph{Proceedings of the 2017 Conference on Conference Human
  Information Interaction and Retrieval}, CHIIR ’17, page 345–348, New
  York, NY, USA. Association for Computing Machinery.

\bibitem[{Buck et~al.(2017)Buck, Bulian, Ciaramita, Gajewski, Gesmundo,
  Houlsby, and Wang}]{buck2017ask}
Christian Buck, Jannis Bulian, Massimiliano Ciaramita, Wojciech Gajewski,
  Andrea Gesmundo, Neil Houlsby, and Wei Wang. 2017.
\newblock \href {http://arxiv.org/abs/1705.07830} {Ask the right questions:
  Active question reformulation with reinforcement learning}.

\bibitem[{Cao et~al.(2008)Cao, Nie, Gao, and
  Robertson}]{10.1145/1390334.1390377}
Guihong Cao, Jian-Yun Nie, Jianfeng Gao, and Stephen Robertson. 2008.
\newblock \href {https://doi.org/10.1145/1390334.1390377} {Selecting good
  expansion terms for pseudo-relevance feedback}.
\newblock In \emph{Proceedings of the 31st Annual International ACM SIGIR
  Conference on Research and Development in Information Retrieval}, SIGIR
  ’08, page 243–250, New York, NY, USA. Association for Computing
  Machinery.

\bibitem[{Celikyilmaz et~al.(2018)Celikyilmaz, Bosselut, He, and
  Choi}]{celikyilmaz-etal-2018-deep}
Asli Celikyilmaz, Antoine Bosselut, Xiaodong He, and Yejin Choi. 2018.
\newblock \href {https://doi.org/10.18653/v1/N18-1150} {Deep communicating
  agents for abstractive summarization}.
\newblock In \emph{Proceedings of the 2018 Conference of the North {A}merican
  Chapter of the Association for Computational Linguistics: Human Language
  Technologies, Volume 1 (Long Papers)}, pages 1662--1675, New Orleans,
  Louisiana. Association for Computational Linguistics.

\bibitem[{Cho et~al.(2019{\natexlab{a}})Cho, Zhang, Zhang, Li, Galley,
  Brockett, Wang, and Gao}]{cho2019towards}
Woon~Sang Cho, Pengchuan Zhang, Yizhe Zhang, Xiujun Li, Michel Galley, Chris
  Brockett, Mengdi Wang, and Jianfeng Gao. 2019{\natexlab{a}}.
\newblock Towards coherent and cohesive long-form text generation.
\newblock In \emph{Proceedings of the First Workshop on Narrative
  Understanding}, pages 1--11.

\bibitem[{Cho et~al.(2019{\natexlab{b}})Cho, Zhang, Rao, Brockett, and
  Lee}]{cho_qgen_2019}
Woon~Sang Cho, Yizhe Zhang, Sudha Rao, Chris Brockett, and Sungjin Lee.
  2019{\natexlab{b}}.
\newblock Generating a common question from multiple documents using
  multi-source encoder-decoder models.
\newblock In \emph{The 3rd Workshop on Neural Generation and Translation}.

\bibitem[{Chu and Liu(2018)}]{DBLP:journals/corr/abs-1810-05739}
Eric Chu and Peter~J. Liu. 2018.
\newblock \href {http://arxiv.org/abs/1810.05739} {Unsupervised neural
  multi-document abstractive summarization}.
\newblock \emph{CoRR}, abs/1810.05739.

\bibitem[{Clark et~al.(2020)Clark, Luong, Le, and Manning}]{clark2020electra}
Kevin Clark, Minh-Thang Luong, Quoc~V. Le, and Christopher~D. Manning. 2020.
\newblock \href {https://openreview.net/pdf?id=r1xMH1BtvB} {{ELECTRA}:
  Pre-training text encoders as discriminators rather than generators}.
\newblock In \emph{ICLR}.

\bibitem[{Collobert et~al.(2011)Collobert, Weston, Bottou, Karlen, Kavukcuoglu,
  and Kuksa}]{JMLR:v12:collobert11a}
Ronan Collobert, Jason Weston, L{{\'e}}on Bottou, Michael Karlen, Koray
  Kavukcuoglu, and Pavel Kuksa. 2011.
\newblock \href {http://jmlr.org/papers/v12/collobert11a.html} {Natural
  language processing (almost) from scratch}.
\newblock \emph{Journal of Machine Learning Research}, 12(76):2493--2537.

\bibitem[{Daum{\'e} et~al.(2009)Daum{\'e}, Langford, and
  Marcu}]{DBLP:journals/corr/abs-0907-0786}
Hal Daum{\'e}, John Langford, and Daniel Marcu. 2009.
\newblock \href {https://doi.org/10.1007/s10994-009-5106-x} {Search-based
  structured prediction}.
\newblock \emph{Machine Learning}, 75(3):297--325.

\bibitem[{Deng et~al.(2020)Deng, Bakhtin, Ott, Szlam, and
  Ranzato}]{Deng2020Residual}
Yuntian Deng, Anton Bakhtin, Myle Ott, Arthur Szlam, and Marc'Aurelio Ranzato.
  2020.
\newblock \href {https://openreview.net/forum?id=B1l4SgHKDH} {Residual
  energy-based models for text generation}.
\newblock In \emph{International Conference on Learning Representations}.

\bibitem[{Dethlefs and Cuay{\'a}huitl(2010)}]{dethlefs2010hierarchical}
Nina Dethlefs and Heriberto Cuay{\'a}huitl. 2010.
\newblock Hierarchical reinforcement learning for adaptive text generation.
\newblock In \emph{Proceedings of the 6th International Natural Language
  Generation Conference}, pages 37--45. Association for Computational
  Linguistics.

\bibitem[{Devlin et~al.(2019)Devlin, Chang, Lee, and
  Toutanova}]{devlin2018bert}
Jacob Devlin, Ming-Wei Chang, Kenton Lee, and Kristina Toutanova. 2019.
\newblock \href {https://doi.org/10.18653/v1/N19-1423} {{BERT}: Pre-training of
  deep bidirectional transformers for language understanding}.
\newblock In \emph{Proceedings of the 2019 Conference of the North {A}merican
  Chapter of the Association for Computational Linguistics: Human Language
  Technologies, Volume 1 (Long and Short Papers)}, pages 4171--4186,
  Minneapolis, Minnesota. Association for Computational Linguistics.

\bibitem[{Dong and Smith(2018)}]{dong-smith-2018-multi}
Rui Dong and David Smith. 2018.
\newblock \href {https://www.aclweb.org/anthology/P18-1220} {Multi-input
  attention for unsupervised {OCR} correction}.
\newblock In \emph{Proceedings of the 56th Annual Meeting of the Association
  for Computational Linguistics (Volume 1: Long Papers)}, pages 2363--2372,
  Melbourne, Australia. Association for Computational Linguistics.

\bibitem[{Dorigo and Colombetti(1994)}]{dorigo1994robot}
Marco Dorigo and Marco Colombetti. 1994.
\newblock Robot shaping: Developing autonomous agents through learning.
\newblock \emph{Artificial intelligence}, 71(2):321--370.

\bibitem[{Du et~al.(2017)Du, Shao, and Cardie}]{du-etal-2017-learning}
Xinya Du, Junru Shao, and Claire Cardie. 2017.
\newblock \href {https://doi.org/10.18653/v1/P17-1123} {Learning to ask: Neural
  question generation for reading comprehension}.
\newblock In \emph{Proceedings of the 55th Annual Meeting of the Association
  for Computational Linguistics (Volume 1: Long Papers)}, pages 1342--1352,
  Vancouver, Canada. Association for Computational Linguistics.

\bibitem[{Duan et~al.(2017)Duan, Tang, Chen, and
  Zhou}]{duan-etal-2017-question}
Nan Duan, Duyu Tang, Peng Chen, and Ming Zhou. 2017.
\newblock \href {https://doi.org/10.18653/v1/D17-1090} {Question generation for
  question answering}.
\newblock In \emph{Proceedings of the 2017 Conference on Empirical Methods in
  Natural Language Processing}, pages 866--874, Copenhagen, Denmark.
  Association for Computational Linguistics.

\bibitem[{Fan et~al.(2018)Fan, Wei, Wang, Liu, and
  Huang}]{fan-etal-2018-reinforcement}
Zhihao Fan, Zhongyu Wei, Siyuan Wang, Yang Liu, and Xuanjing Huang. 2018.
\newblock \href {https://www.aclweb.org/anthology/C18-1150} {A reinforcement
  learning framework for natural question generation using bi-discriminators}.
\newblock In \emph{Proceedings of the 27th International Conference on
  Computational Linguistics}, pages 1763--1774, Santa Fe, New Mexico, USA.
  Association for Computational Linguistics.

\bibitem[{Firat et~al.(2016)Firat, Sankaran, Al{-}Onaizan, Yarman{-}Vural, and
  Cho}]{DBLP:journals/corr/FiratSAYC16}
Orhan Firat, Baskaran Sankaran, Yaser Al{-}Onaizan, Fatos~T. Yarman{-}Vural,
  and Kyunghyun Cho. 2016.
\newblock \href {http://arxiv.org/abs/1606.04164} {Zero-resource translation
  with multi-lingual neural machine translation}.
\newblock \emph{CoRR}, abs/1606.04164.

\bibitem[{Forgues et~al.(2014)Forgues, Pineau, Larchev{\^e}que, and
  Tremblay}]{forgues2014bootstrapping}
Gabriel Forgues, Joelle Pineau, Jean-Marie Larchev{\^e}que, and R{\'e}al
  Tremblay. 2014.
\newblock Bootstrapping dialog systems with word embeddings.
\newblock In \emph{Nips, modern machine learning and natural language
  processing workshop}, volume~2.

\bibitem[{Gao et~al.(2018)Gao, Galley, and Li}]{gaosurvey}
Jianfeng Gao, Michel Galley, and Lihong Li. 2018.
\newblock Neural approaches to conversational {AI}.
\newblock \emph{arXiv preprint arXiv:1809.08267}.

\bibitem[{Garmash and Monz(2016)}]{garmash-monz-2016-ensemble}
Ekaterina Garmash and Christof Monz. 2016.
\newblock \href {https://www.aclweb.org/anthology/C16-1133} {Ensemble learning
  for multi-source neural machine translation}.
\newblock In \emph{Proceedings of {COLING} 2016, the 26th International
  Conference on Computational Linguistics: Technical Papers}, pages 1409--1418,
  Osaka, Japan. The COLING 2016 Organizing Committee.

\bibitem[{Gutmann and Hyvärinen(2010)}]{pmlr-v9-gutmann10a}
Michael Gutmann and Aapo Hyvärinen. 2010.
\newblock \href {http://proceedings.mlr.press/v9/gutmann10a.html}
  {Noise-contrastive estimation: A new estimation principle for unnormalized
  statistical models}.
\newblock In \emph{Proceedings of the Thirteenth International Conference on
  Artificial Intelligence and Statistics}, volume~9 of \emph{Proceedings of
  Machine Learning Research}, pages 297--304, Chia Laguna Resort, Sardinia,
  Italy. PMLR.

\bibitem[{Harrison and Walker(2018)}]{harrison2018neural}
Vrindavan Harrison and Marilyn Walker. 2018.
\newblock \href {http://arxiv.org/abs/1809.02637} {Neural generation of diverse
  questions using answer focus, contextual and linguistic features}.

\bibitem[{Heilman(2011)}]{heilman2011automatic}
Michael Heilman. 2011.
\newblock \emph{Automatic factual question generation from text}.
\newblock Ph.D. thesis, Carnegie Mellon University.

\bibitem[{Hjelm et~al.(2019)Hjelm, Fedorov, Lavoie-Marchildon, Grewal, Bachman,
  Trischler, and Bengio}]{hjelm2019learning}
Devon Hjelm, Alex Fedorov, Samuel Lavoie-Marchildon, Karan Grewal, Philip
  Bachman, Adam Trischler, and Yoshua Bengio. 2019.
\newblock \href
  {https://www.microsoft.com/en-us/research/publication/learning-deep-representations-by-mutual-information-estimation-and-maximization/}
  {Learning deep representations by mutual information estimation and
  maximization}.
\newblock In \emph{ICLR 2019}. ICLR.

\bibitem[{Hochreiter and
  Schmidhuber(1997)}]{Hochreiter:1997:LSM:1246443.1246450}
Sepp Hochreiter and J\"{u}rgen Schmidhuber. 1997.
\newblock \href {https://doi.org/10.1162/neco.1997.9.8.1735} {Long short-term
  memory}.
\newblock \emph{Neural Comput.}, 9(8):1735--1780.

\bibitem[{Janarthanam and Lemon(2009)}]{janarthanam-lemon-2009-learning}
Srinivasan Janarthanam and Oliver Lemon. 2009.
\newblock \href {https://www.aclweb.org/anthology/W09-0611} {Learning lexical
  alignment policies for generating referring expressions for spoken dialogue
  systems}.
\newblock In \emph{Proceedings of the 12th {E}uropean Workshop on Natural
  Language Generation ({ENLG} 2009)}, pages 74--81, Athens, Greece. Association
  for Computational Linguistics.

\bibitem[{Jaques et~al.(2017)Jaques, Gu, Bahdanau, Hern{\'a}ndez-Lobato,
  Turner, and Eck}]{jaques2017sequence}
Natasha Jaques, Shixiang Gu, Dzmitry Bahdanau, Jos{\'e}~Miguel
  Hern{\'a}ndez-Lobato, Richard~E Turner, and Douglas Eck. 2017.
\newblock Sequence tutor: Conservative fine-tuning of sequence generation
  models with kl-control.
\newblock In \emph{Proceedings of the 34th International Conference on Machine
  Learning-Volume 70}, pages 1645--1654. JMLR. org.

\bibitem[{J\"{a}rvelin and
  Kek\"{a}l\"{a}inen(2002)}]{Jarvelin:2002:CGE:582415.582418}
Kalervo J\"{a}rvelin and Jaana Kek\"{a}l\"{a}inen. 2002.
\newblock \href {https://doi.org/10.1145/582415.582418} {Cumulated gain-based
  evaluation of ir techniques}.
\newblock \emph{ACM Trans. Inf. Syst.}, 20(4):422--446.

\bibitem[{Jones(1972)}]{Jones72astatistical}
Karen~Spärck Jones. 1972.
\newblock A statistical interpretation of term specificity and its application
  in retrieval.
\newblock \emph{Journal of Documentation}, 28:11--21.

\bibitem[{Kalchbrenner and Blunsom(2013)}]{kalchbrenner2013recurrent}
Nal Kalchbrenner and Phil Blunsom. 2013.
\newblock Recurrent continuous translation models.
\newblock In \emph{Proceedings of the 2013 Conference on Empirical Methods in
  Natural Language Processing}, pages 1700--1709.

\bibitem[{Kenter et~al.(2016)Kenter, Borisov, and De~Rijke}]{kenter2016siamese}
Tom Kenter, Alexey Borisov, and Maarten De~Rijke. 2016.
\newblock Siamese cbow: Optimizing word embeddings for sentence
  representations.
\newblock \emph{arXiv preprint arXiv:1606.04640}.

\bibitem[{Kingma and Ba(2014)}]{Kingma2014AdamAM}
Diederik~P. Kingma and Jimmy Ba. 2014.
\newblock Adam: A method for stochastic optimization.
\newblock \emph{CoRR}, abs/1412.6980.

\bibitem[{Kiros et~al.(2015)Kiros, Zhu, Salakhutdinov, Zemel, Torralba,
  Urtasun, and Fidler}]{kiros2015skip}
Ryan Kiros, Yukun Zhu, Ruslan Salakhutdinov, Richard~S Zemel, Antonio Torralba,
  Raquel Urtasun, and Sanja Fidler. 2015.
\newblock Skip-thought vectors.
\newblock \emph{arXiv preprint arXiv:1506.06726}.

\bibitem[{Labutov et~al.(2015)Labutov, Basu, and
  Vanderwende}]{labutov-etal-2015-deep}
Igor Labutov, Sumit Basu, and Lucy Vanderwende. 2015.
\newblock \href {https://doi.org/10.3115/v1/P15-1086} {Deep questions without
  deep understanding}.
\newblock In \emph{Proceedings of the 53rd Annual Meeting of the Association
  for Computational Linguistics and the 7th International Joint Conference on
  Natural Language Processing (Volume 1: Long Papers)}, pages 889--898,
  Beijing, China. Association for Computational Linguistics.

\bibitem[{Lebanoff et~al.(2018)Lebanoff, Song, and
  Liu}]{lebanoff-etal-2018-adapting}
Logan Lebanoff, Kaiqiang Song, and Fei Liu. 2018.
\newblock \href {https://www.aclweb.org/anthology/D18-1446} {Adapting the
  neural encoder-decoder framework from single to multi-document
  summarization}.
\newblock In \emph{Proceedings of the 2018 Conference on Empirical Methods in
  Natural Language Processing}, pages 4131--4141, Brussels, Belgium.
  Association for Computational Linguistics.

\bibitem[{Lemon(2008)}]{lemon2008adaptive}
Oliver Lemon. 2008.
\newblock Adaptive natural language generation in dialogue using reinforcement
  learning.
\newblock In \emph{Proceedings of the 12th SEMdial Workshop on the Semantics
  and Pragmatics of Dialogues}, pages 149--156.

\bibitem[{Lewis et~al.(2017)Lewis, Yarats, Dauphin, Parikh, and
  Batra}]{lewis2017deal}
Mike Lewis, Denis Yarats, Yann~N Dauphin, Devi Parikh, and Dhruv Batra. 2017.
\newblock Deal or no deal? end-to-end learning for negotiation dialogues.
\newblock \emph{arXiv preprint arXiv:1706.05125}.

\bibitem[{Li et~al.(2018{\natexlab{a}})Li, Sun, He, Wang, Hui, Yates, Sun, and
  Xu}]{li-etal-2018-nprf}
Canjia Li, Yingfei Sun, Ben He, Le~Wang, Kai Hui, Andrew Yates, Le~Sun, and
  Jungang Xu. 2018{\natexlab{a}}.
\newblock \href {https://www.aclweb.org/anthology/D18-1478} {{NPRF}: A neural
  pseudo relevance feedback framework for ad-hoc information retrieval}.
\newblock In \emph{Proceedings of the 2018 Conference on Empirical Methods in
  Natural Language Processing}, pages 4482--4491, Brussels, Belgium.
  Association for Computational Linguistics.

\bibitem[{Li et~al.(2018{\natexlab{b}})Li, Wang, Mallidi, Hori, Watanabe, and
  Hermansky}]{DBLP:journals/corr/abs-1811-04897}
Ruizhi Li, Xiaofei Wang, Sri Harish~Reddy Mallidi, Takaaki Hori, Shinji
  Watanabe, and Hynek Hermansky. 2018{\natexlab{b}}.
\newblock \href {http://arxiv.org/abs/1811.04897} {Multi-encoder
  multi-resolution framework for end-to-end speech recognition}.
\newblock \emph{CoRR}, abs/1811.04897.

\bibitem[{Libovick{\'y} and Helcl(2017)}]{libovicky-helcl-2017-attention}
Jind{\v{r}}ich Libovick{\'y} and Jind{\v{r}}ich Helcl. 2017.
\newblock \href {https://doi.org/10.18653/v1/P17-2031} {Attention strategies
  for multi-source sequence-to-sequence learning}.
\newblock In \emph{Proceedings of the 55th Annual Meeting of the Association
  for Computational Linguistics (Volume 2: Short Papers)}, pages 196--202,
  Vancouver, Canada. Association for Computational Linguistics.

\bibitem[{Libovick{\'y} et~al.(2018)Libovick{\'y}, Helcl, and
  Mare{\v{c}}ek}]{libovicky-etal-2018-input}
Jind{\v{r}}ich Libovick{\'y}, Jind{\v{r}}ich Helcl, and David Mare{\v{c}}ek.
  2018.
\newblock \href {https://www.aclweb.org/anthology/W18-6326} {Input combination
  strategies for multi-source transformer decoder}.
\newblock In \emph{Proceedings of the Third Conference on Machine Translation:
  Research Papers}, pages 253--260, Belgium, Brussels. Association for
  Computational Linguistics.

\bibitem[{Lin and Hovy(2003)}]{Lin:2003:AES:1073445.1073465}
Chin-Yew Lin and Eduard Hovy. 2003.
\newblock Automatic evaluation of summaries using n-gram co-occurrence
  statistics.
\newblock In \emph{Proceedings of the 2003 Conference of the North American
  Chapter of the Association for Computational Linguistics on Human Language
  Technology - Volume 1}, NAACL '03, pages 71--78, Stroudsburg, PA, USA.

\bibitem[{Liu et~al.(2018)Liu, Saleh, Pot, Goodrich, Sepassi, Kaiser, and
  Shazeer}]{liu2018generating}
Peter~J Liu, Mohammad Saleh, Etienne Pot, Ben Goodrich, Ryan Sepassi, Lukasz
  Kaiser, and Noam Shazeer. 2018.
\newblock Generating wikipedia by summarizing long sequences.
\newblock \emph{International Conference on Learning Representations}.

\bibitem[{Liu and Lapata(2019)}]{lapata2019}
Yang Liu and Mirala Lapata. 2019.
\newblock Hierarchical transformers for multidocument summarization.
\newblock In \emph{ACL}.

\bibitem[{Luan et~al.(2016)Luan, Ji, Hajishirzi, and
  Li}]{luan2016multiplicative}
Yi~Luan, Yangfeng Ji, Hannaneh Hajishirzi, and Boyang Li. 2016.
\newblock Multiplicative representations for unsupervised semantic role
  induction.
\newblock In \emph{ACL}.

\bibitem[{Luhn(1957)}]{luhn1957statistical}
Hans~Peter Luhn. 1957.
\newblock A statistical approach to mechanized encoding and searching of
  literary information.
\newblock \emph{IBM Journal of research and development}, 1(4):309--317.

\bibitem[{Luong et~al.(2015)Luong, Pham, and
  Manning}]{luong-etal-2015-effective}
Thang Luong, Hieu Pham, and Christopher~D. Manning. 2015.
\newblock \href {https://doi.org/10.18653/v1/D15-1166} {Effective approaches to
  attention-based neural machine translation}.
\newblock In \emph{Proceedings of the 2015 Conference on Empirical Methods in
  Natural Language Processing}, pages 1412--1421, Lisbon, Portugal. Association
  for Computational Linguistics.

\bibitem[{Metzler and Croft(2007)}]{Metzler:2007:LCE:1277741.1277796}
Donald Metzler and W.~Bruce Croft. 2007.
\newblock \href {https://doi.org/10.1145/1277741.1277796} {Latent concept
  expansion using markov random fields}.
\newblock In \emph{Proceedings of the 30th Annual International ACM SIGIR
  Conference on Research and Development in Information Retrieval}, SIGIR '07,
  pages 311--318, New York, NY, USA. ACM.

\bibitem[{Mikolov et~al.(2013)Mikolov, Sutskever, Chen, Corrado, and
  Dean}]{mikolov2013distributed}
Tomas Mikolov, Ilya Sutskever, Kai Chen, Greg~S Corrado, and Jeff Dean. 2013.
\newblock Distributed representations of words and phrases and their
  compositionality.
\newblock In \emph{Advances in neural information processing systems}, pages
  3111--3119.

\bibitem[{Mostafazadeh et~al.(2017)Mostafazadeh, Brockett, Dolan, Galley, Gao,
  Spithourakis, and Vanderwende}]{mostafazadeh-etal-2017-image}
Nasrin Mostafazadeh, Chris Brockett, Bill Dolan, Michel Galley, Jianfeng Gao,
  Georgios Spithourakis, and Lucy Vanderwende. 2017.
\newblock \href {https://www.aclweb.org/anthology/I17-1047} {Image-grounded
  conversations: Multimodal context for natural question and response
  generation}.
\newblock In \emph{Proceedings of the Eighth International Joint Conference on
  Natural Language Processing (Volume 1: Long Papers)}, pages 462--472, Taipei,
  Taiwan. Asian Federation of Natural Language Processing.

\bibitem[{Nair and Hinton(2010)}]{Nair:2010:RLU:3104322.3104425}
Vinod Nair and Geoffrey~E. Hinton. 2010.
\newblock \href {http://dl.acm.org/citation.cfm?id=3104322.3104425} {Rectified
  linear units improve restricted boltzmann machines}.
\newblock In \emph{Proceedings of the 27th International Conference on
  International Conference on Machine Learning}, ICML'10, pages 807--814, USA.
  Omnipress.

\bibitem[{Ng et~al.(1999)Ng, Harada, and Russell}]{ng1999policy}
Andrew~Y Ng, Daishi Harada, and Stuart Russell. 1999.
\newblock Policy invariance under reward transformations: Theory and
  application to reward shaping.
\newblock In \emph{ICML}, volume~99, pages 278--287.

\bibitem[{Nguyen et~al.(2016)Nguyen, Rosenberg, Song, Gao, Tiwary, Majumder,
  and Deng}]{DBLP:conf/nips/NguyenRSGTMD16}
Tri Nguyen, Mir Rosenberg, Xia Song, Jianfeng Gao, Saurabh Tiwary, Rangan
  Majumder, and Li~Deng. 2016.
\newblock \href {http://ceur-ws.org/Vol-1773/CoCoNIPS\_2016\_paper9.pdf} {{MS}
  {MARCO:} {A} human generated machine reading comprehension dataset}.
\newblock In \emph{Proceedings of the Workshop on Cognitive Computation:
  Integrating neural and symbolic approaches 2016 co-located with the 30th
  Annual Conference on Neural Information Processing Systems {(NIPS} 2016),
  Barcelona, Spain, December 9, 2016.}

\bibitem[{Nishida et~al.(2019)Nishida, Saito, Nishida, Shinoda, Otsuka, Asano,
  and Tomita}]{DBLP:journals/corr/abs-1901-02262}
Kyosuke Nishida, Itsumi Saito, Kosuke Nishida, Kazutoshi Shinoda, Atsushi
  Otsuka, Hisako Asano, and Junji Tomita. 2019.
\newblock \href {http://arxiv.org/abs/1901.02262} {Multi-style generative
  reading comprehension}.
\newblock \emph{CoRR}, abs/1901.02262.

\bibitem[{Nishimura et~al.(2018)Nishimura, Sudoh, Neubig, and
  Nakamura}]{nishimura-etal-2018-multi}
Yuta Nishimura, Katsuhito Sudoh, Graham Neubig, and Satoshi Nakamura. 2018.
\newblock \href {https://www.aclweb.org/anthology/W18-2711} {Multi-source
  neural machine translation with missing data}.
\newblock In \emph{Proceedings of the 2nd Workshop on Neural Machine
  Translation and Generation}, pages 92--99, Melbourne, Australia. Association
  for Computational Linguistics.

\bibitem[{Nogueira and Cho(2017)}]{nogueira-cho-2017-task}
Rodrigo Nogueira and Kyunghyun Cho. 2017.
\newblock \href {https://doi.org/10.18653/v1/D17-1061} {Task-oriented query
  reformulation with reinforcement learning}.
\newblock In \emph{Proceedings of the 2017 Conference on Empirical Methods in
  Natural Language Processing}, pages 574--583, Copenhagen, Denmark.
  Association for Computational Linguistics.

\bibitem[{Nogueira and Cho(2019)}]{DBLP:journals/corr/abs-1901-04085}
Rodrigo Nogueira and Kyunghyun Cho. 2019.
\newblock \href {http://arxiv.org/abs/1901.04085} {Passage re-ranking with
  {BERT}}.
\newblock \emph{CoRR}, abs/1901.04085.

\bibitem[{Och and Ney(2001)}]{Och01statisticalmulti-source}
Franz~Josef Och and Hermann Ney. 2001.
\newblock Statistical multi-source translation.
\newblock In \emph{In MT Summit 2001}, pages 253--258.

\bibitem[{Pan et~al.(2019)Pan, Lei, Chua, and
  Kan}]{DBLP:journals/corr/abs-1905-08949}
Liangming Pan, Wenqiang Lei, Tat{-}Seng Chua, and Min{-}Yen Kan. 2019.
\newblock \href {http://arxiv.org/abs/1905.08949} {Recent advances in neural
  question generation}.
\newblock \emph{CoRR}, abs/1905.08949.

\bibitem[{Papineni et~al.(2002)Papineni, Roukos, Ward, and
  Zhu}]{papineni2002bleu}
Kishore Papineni, Salim Roukos, Todd Ward, and Wei-Jing Zhu. 2002.
\newblock Bleu: a method for automatic evaluation of machine translation.
\newblock In \emph{Proceedings of the 40th annual meeting on association for
  computational linguistics}, pages 311--318. Association for Computational
  Linguistics.

\bibitem[{Parveen and Strube(2014)}]{dragomir}
Daraksha Parveen and Michael Strube. 2014.
\newblock Multidocument summarization using bipartite graphs.
\newblock In \emph{TextGraphs-9: the workshop on Graph-based Methods for
  Natural Language Processing}.

\bibitem[{Pasunuru and Bansal(2017)}]{pasunuru2017reinforced}
Ramakanth Pasunuru and Mohit Bansal. 2017.
\newblock Reinforced video captioning with entailment rewards.
\newblock \emph{arXiv preprint arXiv:1708.02300}.

\bibitem[{Paulus et~al.(2017)Paulus, Xiong, and
  Socher}]{DBLP:journals/corr/PaulusXS17}
Romain Paulus, Caiming Xiong, and Richard Socher. 2017.
\newblock \href {http://arxiv.org/abs/1705.04304} {A deep reinforced model for
  abstractive summarization}.
\newblock \emph{CoRR}, abs/1705.04304.

\bibitem[{Pennington et~al.(2014)Pennington, Socher, and
  Manning}]{pennington2014glove}
Jeffrey Pennington, Richard Socher, and Christopher~D. Manning. 2014.
\newblock {GloVe}: Global vectors for word representation.
\newblock In \emph{Proceedings of the 2014 Conference on Empirical Methods in
  Natural Language Processing}, pages 1532--1543.

\bibitem[{Peters et~al.(2018)Peters, Neumann, Iyyer, Gardner, Clark, Lee, and
  Zettlemoyer}]{Peters:2018}
Matthew~E. Peters, Mark Neumann, Mohit Iyyer, Matt Gardner, Christopher Clark,
  Kenton Lee, and Luke Zettlemoyer. 2018.
\newblock Deep contextualized word representations.
\newblock In \emph{Proc. of NAACL}.

\bibitem[{Qi et~al.(2020)Qi, Zhang, and Manning}]{qi2020stay}
Peng Qi, Yuhao Zhang, and Christopher~D Manning. 2020.
\newblock Stay hungry, stay focused: Generating informative and specific
  questions in information-seeking conversations.
\newblock \emph{arXiv preprint arXiv:2004.14530}.

\bibitem[{Radev(2000)}]{commondragomir}
Dragomir Radev. 2000.
\newblock A common theory of information fusion from multiple text sources step
  one: Cross-document structure.
\newblock In \emph{1st SIGdial Workshop on Discourse and Dialogue}, pages
  78--83.

\bibitem[{Radev et~al.(2002)Radev, Qi, Wu, and
  Fan}]{radev-etal-2002-evaluating}
Dragomir~R. Radev, Hong Qi, Harris Wu, and Weiguo Fan. 2002.
\newblock \href {http://www.lrec-conf.org/proceedings/lrec2002/pdf/301.pdf}
  {Evaluating web-based question answering systems}.
\newblock In \emph{Proceedings of the Third International Conference on
  Language Resources and Evaluation ({LREC}{'}02)}, Las Palmas, Canary Islands
  - Spain. European Language Resources Association (ELRA).

\bibitem[{Radford et~al.(2019)Radford, Wu, Child, Luan, Amodei, and
  Sutskever}]{radford2019language}
Alec Radford, Jeff Wu, Rewon Child, David Luan, Dario Amodei, and Ilya
  Sutskever. 2019.
\newblock Language models are unsupervised multitask learners.

\bibitem[{Radlinski and Craswell(2017)}]{10.1145/3020165.3020183}
Filip Radlinski and Nick Craswell. 2017.
\newblock \href {https://doi.org/10.1145/3020165.3020183} {A theoretical
  framework for conversational search}.
\newblock In \emph{Proceedings of the 2017 Conference on Conference Human
  Information Interaction and Retrieval}, CHIIR ’17, page 117–126, New
  York, NY, USA. Association for Computing Machinery.

\bibitem[{Raffel et~al.(2019)Raffel, Shazeer, Roberts, Lee, Narang, Matena,
  Zhou, Li, and Liu}]{2019t5}
Colin Raffel, Noam Shazeer, Adam Roberts, Katherine Lee, Sharan Narang, Michael
  Matena, Yanqi Zhou, Wei Li, and Peter~J. Liu. 2019.
\newblock Exploring the limits of transfer learning with a unified text-to-text
  transformer.
\newblock \emph{arXiv e-prints}.

\bibitem[{Ranzato et~al.(2015)Ranzato, Chopra, Auli, and
  Zaremba}]{DBLP:journals/corr/RanzatoCAZ15}
Marc'Aurelio Ranzato, Sumit Chopra, Michael Auli, and Wojciech Zaremba. 2015.
\newblock \href {http://arxiv.org/abs/1511.06732} {Sequence level training with
  recurrent neural networks}.
\newblock \emph{CoRR}, abs/1511.06732.

\bibitem[{Rao and Daum{\'e}~III(2018)}]{rao-daume-iii-2018-learning}
Sudha Rao and Hal Daum{\'e}~III. 2018.
\newblock \href {https://doi.org/10.18653/v1/P18-1255} {Learning to ask good
  questions: Ranking clarification questions using neural expected value of
  perfect information}.
\newblock In \emph{Proceedings of the 56th Annual Meeting of the Association
  for Computational Linguistics (Volume 1: Long Papers)}, pages 2737--2746,
  Melbourne, Australia. Association for Computational Linguistics.

\bibitem[{Rao and Daum{\'e}~III(2019)}]{rao-daume-iii-2019-answer}
Sudha Rao and Hal Daum{\'e}~III. 2019.
\newblock \href {https://doi.org/10.18653/v1/N19-1013} {{A}nswer-based
  {A}dversarial {T}raining for {G}enerating {C}larification {Q}uestions}.
\newblock In \emph{Proceedings of the 2019 Conference of the North {A}merican
  Chapter of the Association for Computational Linguistics: Human Language
  Technologies, Volume 1 (Long and Short Papers)}, pages 143--155, Minneapolis,
  Minnesota. Association for Computational Linguistics.

\bibitem[{Rennie et~al.(2017)Rennie, Marcheret, Mroueh, Ross, and
  Goel}]{rennie2017self}
Steven~J Rennie, Etienne Marcheret, Youssef Mroueh, Jerret Ross, and Vaibhava
  Goel. 2017.
\newblock Self-critical sequence training for image captioning.
\newblock In \emph{Proceedings of the IEEE Conference on Computer Vision and
  Pattern Recognition}, pages 7008--7024.

\bibitem[{Rieser and Lemon(2009)}]{rieser2009natural}
Verena Rieser and Oliver Lemon. 2009.
\newblock Natural language generation as planning under uncertainty for spoken
  dialogue systems.
\newblock In \emph{Empirical methods in natural language generation}, pages
  105--120. Springer.

\bibitem[{Robertson and Zaragoza(2009)}]{Robertson:2009:PRF:1704809.1704810}
Stephen Robertson and Hugo Zaragoza. 2009.
\newblock \href {https://doi.org/10.1561/1500000019} {The probabilistic
  relevance framework: Bm25 and beyond}.
\newblock \emph{Found. Trends Inf. Retr.}, 3(4):333--389.

\bibitem[{Rocchio(1971)}]{rocchio1971relevance}
Joseph Rocchio. 1971.
\newblock Relevance feedback in information retrieval.
\newblock \emph{The Smart retrieval system-experiments in automatic document
  processing}, pages 313--323.

\bibitem[{Rokach(2010)}]{Rokach2010}
Lior Rokach. 2010.
\newblock \href {https://doi.org/10.1007/s10462-009-9124-7} {Ensemble-based
  classifiers}.
\newblock \emph{Artificial Intelligence Review}, 33(1):1--39.

\bibitem[{Rothe et~al.(2016)Rothe, Lake, and
  Gureckis}]{a92dd6bf37bd48f99a691c3ea8343823}
A~Rothe, Brenden Lake, and Todd Gureckis. 2016.
\newblock Asking and evaluating natural language questions.
\newblock In \emph{Proceedings of the 38th Annual Conference of the Cognitive
  Science Society}.

\bibitem[{Rumelhart et~al.(1988)Rumelhart, Hinton, Williams
  et~al.}]{rumelhart1988learning}
David~E Rumelhart, Geoffrey~E Hinton, Ronald~J Williams, et~al. 1988.
\newblock Learning representations by back-propagating errors.
\newblock \emph{Cognitive modeling}, 5(3):1.

\bibitem[{Rus and Lintean(2012)}]{rus2012comparison}
Vasile Rus and Mihai Lintean. 2012.
\newblock A comparison of greedy and optimal assessment of natural language
  student input using word-to-word similarity metrics.
\newblock In \emph{Proceedings of the Seventh Workshop on Building Educational
  Applications Using NLP}, pages 157--162. Association for Computational
  Linguistics.

\bibitem[{Rus et~al.(2010)Rus, Wyse, Piwek, Lintean, Stoyanchev, and
  Moldovan}]{rus2010first}
Vasile Rus, Brendan Wyse, Paul Piwek, Mihai Lintean, Svetlana Stoyanchev, and
  Cristian Moldovan. 2010.
\newblock The first question generation shared task evaluation challenge.
\newblock In \emph{Proceedings of the 6th International Natural Language
  Generation Conference}, pages 251--257. Association for Computational
  Linguistics.

\bibitem[{Salton(1971)}]{Salton:1971:SRS:1102022}
G.~Salton. 1971.
\newblock \emph{The SMART Retrieval System---Experiments in Automatic Document
  Processing}.
\newblock Prentice-Hall, Inc., Upper Saddle River, NJ, USA.

\bibitem[{Salton and McGill(1983)}]{salton1983introduction}
Gerard Salton and Michael~J McGill. 1983.
\newblock \emph{Introduction to modern information retrieval}.
\newblock mcgraw-hill.

\bibitem[{See et~al.(2017)See, Liu, and Manning}]{see-etal-2017-get}
Abigail See, Peter~J. Liu, and Christopher~D. Manning. 2017.
\newblock \href {https://doi.org/10.18653/v1/P17-1099} {Get to the point:
  Summarization with pointer-generator networks}.
\newblock In \emph{Proceedings of the 55th Annual Meeting of the Association
  for Computational Linguistics (Volume 1: Long Papers)}, pages 1073--1083,
  Vancouver, Canada. Association for Computational Linguistics.

\bibitem[{Serban et~al.(2016)Serban, Garc{\'\i}a-Dur{\'a}n, Gulcehre, Ahn,
  Chandar, Courville, and Bengio}]{serban-etal-2016-generating}
Iulian~Vlad Serban, Alberto Garc{\'\i}a-Dur{\'a}n, Caglar Gulcehre, Sungjin
  Ahn, Sarath Chandar, Aaron Courville, and Yoshua Bengio. 2016.
\newblock \href {https://doi.org/10.18653/v1/P16-1056} {Generating factoid
  questions with recurrent neural networks: The 30{M} factoid question-answer
  corpus}.
\newblock In \emph{Proceedings of the 54th Annual Meeting of the Association
  for Computational Linguistics (Volume 1: Long Papers)}, pages 588--598,
  Berlin, Germany. Association for Computational Linguistics.

\bibitem[{Sharma et~al.(2017)Sharma, El~Asri, Schulz, and
  Zumer}]{sharma2017nlgeval}
Shikhar Sharma, Layla El~Asri, Hannes Schulz, and Jeremie Zumer. 2017.
\newblock \href {http://arxiv.org/abs/1706.09799} {Relevance of unsupervised
  metrics in task-oriented dialogue for evaluating natural language
  generation}.
\newblock \emph{CoRR}, abs/1706.09799.

\bibitem[{Smith and Eisner(2005)}]{smith-eisner-2005-contrastive}
Noah~A. Smith and Jason Eisner. 2005.
\newblock \href {https://doi.org/10.3115/1219840.1219884} {Contrastive
  estimation: Training log-linear models on unlabeled data}.
\newblock In \emph{Proceedings of the 43rd Annual Meeting of the Association
  for Computational Linguistics ({ACL}{'}05)}, pages 354--362, Ann Arbor,
  Michigan. Association for Computational Linguistics.

\bibitem[{Song et~al.(2017)Song, Wang, and Hamza}]{song2017unified}
Linfeng Song, Zhiguo Wang, and Wael Hamza. 2017.
\newblock \href {http://arxiv.org/abs/1709.01058} {A unified query-based
  generative model for question generation and question answering}.

\bibitem[{Song et~al.(2018)Song, Wang, Hamza, Zhang, and
  Gildea}]{song-etal-2018-leveraging}
Linfeng Song, Zhiguo Wang, Wael Hamza, Yue Zhang, and Daniel Gildea. 2018.
\newblock \href {https://doi.org/10.18653/v1/N18-2090} {Leveraging context
  information for natural question generation}.
\newblock In \emph{Proceedings of the 2018 Conference of the North {A}merican
  Chapter of the Association for Computational Linguistics: Human Language
  Technologies, Volume 2 (Short Papers)}, pages 569--574, New Orleans,
  Louisiana. Association for Computational Linguistics.

\bibitem[{Sun et~al.(2018)Sun, Liu, Lyu, He, Ma, and
  Wang}]{sun-etal-2018-answer}
Xingwu Sun, Jing Liu, Yajuan Lyu, Wei He, Yanjun Ma, and Shi Wang. 2018.
\newblock \href {https://www.aclweb.org/anthology/D18-1427} {Answer-focused and
  position-aware neural question generation}.
\newblock In \emph{Proceedings of the 2018 Conference on Empirical Methods in
  Natural Language Processing}, pages 3930--3939, Brussels, Belgium.
  Association for Computational Linguistics.

\bibitem[{Sutskever et~al.(2014)Sutskever, Vinyals, and Le}]{seq2seq}
Ilya Sutskever, Oriol Vinyals, and Quoc~V Le. 2014.
\newblock Sequence to sequence learning with neural networks.
\newblock In \emph{NIPS}.

\bibitem[{Sutton et~al.(1999)Sutton, McAllester, Singh, and
  Mansour}]{Sutton:1999:PGM:3009657.3009806}
Richard~S. Sutton, David McAllester, Satinder Singh, and Yishay Mansour. 1999.
\newblock Policy gradient methods for reinforcement learning with function
  approximation.
\newblock In \emph{Proceedings of the 12th International Conference on Neural
  Information Processing Systems}, NIPS'99, pages 1057--1063. MIT Press.

\bibitem[{Tu et~al.(2016)Tu, Lu, Liu, Liu, and Li}]{tu-etal-2016-modeling}
Zhaopeng Tu, Zhengdong Lu, Yang Liu, Xiaohua Liu, and Hang Li. 2016.
\newblock \href {https://doi.org/10.18653/v1/P16-1008} {Modeling coverage for
  neural machine translation}.
\newblock In \emph{Proceedings of the 54th Annual Meeting of the Association
  for Computational Linguistics (Volume 1: Long Papers)}, pages 76--85, Berlin,
  Germany. Association for Computational Linguistics.

\bibitem[{Vaswani et~al.(2017)Vaswani, Shazeer, Parmar, Uszkoreit, Jones,
  Gomez, Kaiser, and Polosukhin}]{vaswani2017attention}
Ashish Vaswani, Noam Shazeer, Niki Parmar, Jakob Uszkoreit, Llion Jones,
  Aidan~N Gomez, {\L}ukasz Kaiser, and Illia Polosukhin. 2017.
\newblock Attention is all you need.
\newblock In \emph{Advances in neural information processing systems}, pages
  5998--6008.

\bibitem[{Vedantam et~al.(2015)Vedantam, Lawrence~Zitnick, and
  Parikh}]{vedantam2015cider}
Ramakrishna Vedantam, C~Lawrence~Zitnick, and Devi Parikh. 2015.
\newblock Cider: Consensus-based image description evaluation.
\newblock In \emph{Proceedings of the IEEE conference on computer vision and
  pattern recognition}, pages 4566--4575.

\bibitem[{Voorhees(2001)}]{Voorhees:2001:TQA:973890.973895}
Ellen~M. Voorhees. 2001.
\newblock \href {https://doi.org/10.1017/S1351324901002789} {The trec question
  answering track}.
\newblock \emph{Nat. Lang. Eng.}, 7(4):361--378.

\bibitem[{Voorhees(1999)}]{voorhees1999proceedings}
EM~Voorhees. 1999.
\newblock Proceedings of the 8th text retrieval conference.
\newblock \emph{TREC-8 Question Answering Track Report}, pages 77--82.

\bibitem[{Wan(2008)}]{wan2008}
Xiaojun Wan. 2008.
\newblock An exploration of document impact on graph-based multi-document
  summarization.
\newblock In \emph{Conference on Empirical Methods in Natural Language
  Processing}.

\bibitem[{Wang et~al.(2018)Wang, Liu, Liu, He, Lyu, Wu, Li, and
  Wang}]{wang-etal-2018-multi-passage}
Yizhong Wang, Kai Liu, Jing Liu, Wei He, Yajuan Lyu, Hua Wu, Sujian Li, and
  Haifeng Wang. 2018.
\newblock \href {https://www.aclweb.org/anthology/P18-1178} {Multi-passage
  machine reading comprehension with cross-passage answer verification}.
\newblock In \emph{Proceedings of the 56th Annual Meeting of the Association
  for Computational Linguistics (Volume 1: Long Papers)}, pages 1918--1927,
  Melbourne, Australia. Association for Computational Linguistics.

\bibitem[{{Werbos}(1990)}]{58337}
P.~J. {Werbos}. 1990.
\newblock \href {https://doi.org/10.1109/5.58337} {Backpropagation through
  time: what it does and how to do it}.
\newblock \emph{Proceedings of the IEEE}, 78(10):1550--1560.

\bibitem[{Wu et~al.(2016)Wu, Schuster, Chen, Le, Norouzi, Macherey, Krikun,
  Cao, Gao, Macherey et~al.}]{wu2016google}
Yonghui Wu, Mike Schuster, Zhifeng Chen, Quoc~V Le, Mohammad Norouzi, Wolfgang
  Macherey, Maxim Krikun, Yuan Cao, Qin Gao, Klaus Macherey, et~al. 2016.
\newblock Google's neural machine translation system: Bridging the gap between
  human and machine translation.
\newblock \emph{arXiv preprint arXiv:1609.08144}.

\bibitem[{Xu and Croft(1996)}]{Xu:1996:QEU:243199.243202}
Jinxi Xu and W.~Bruce Croft. 1996.
\newblock \href {https://doi.org/10.1145/243199.243202} {Query expansion using
  local and global document analysis}.
\newblock In \emph{Proceedings of the 19th Annual International ACM SIGIR
  Conference on Research and Development in Information Retrieval}, SIGIR '96,
  pages 4--11, New York, NY, USA. ACM.

\bibitem[{Yan et~al.(2018)Yan, Xia, Wu, Bi, Zhao, Zhang, Si, Wang, Wang, and
  Chen}]{DBLP:journals/corr/abs-1811-11374}
Ming Yan, Jiangnan Xia, Chen Wu, Bin Bi, Zhongzhou Zhao, Ji~Zhang, Luo Si, Rui
  Wang, Wei Wang, and Haiqing Chen. 2018.
\newblock \href {http://arxiv.org/abs/1811.11374} {A deep cascade model for
  multi-document reading comprehension}.
\newblock \emph{CoRR}, abs/1811.11374.

\bibitem[{Zamani et~al.(2020)Zamani, Dumais, Craswell, Bennett, and
  Lueck}]{zamani2020generating}
Hamed Zamani, Susan Dumais, Nick Craswell, Paul Bennett, and Gord Lueck. 2020.
\newblock \href
  {https://www.microsoft.com/en-us/research/publication/generating-clarifying-questions-for-information-retrieval/}
  {Generating clarifying questions for information retrieval}.
\newblock In \emph{The Web Conference 2020 (formerly WWW conference)}.

\bibitem[{Zhai and Lafferty(2001)}]{Zhai:2001:MFL:502585.502654}
Chengxiang Zhai and John Lafferty. 2001.
\newblock \href {https://doi.org/10.1145/502585.502654} {Model-based feedback
  in the language modeling approach to information retrieval}.
\newblock In \emph{Proceedings of the Tenth International Conference on
  Information and Knowledge Management}, CIKM '01, pages 403--410, New York,
  NY, USA. ACM.

\bibitem[{Zhang et~al.(2019)Zhang, Song, Xiong, Rosset, Bennett, Craswell, and
  Tiwary}]{Zhang:2019:GIR:3331184.3331198}
Hongfei Zhang, Xia Song, Chenyan Xiong, Corby Rosset, Paul~N. Bennett, Nick
  Craswell, and Saurabh Tiwary. 2019.
\newblock \href {https://doi.org/10.1145/3331184.3331198} {Generic intent
  representation in web search}.
\newblock In \emph{Proceedings of the 42Nd International ACM SIGIR Conference
  on Research and Development in Information Retrieval}, SIGIR'19, pages
  65--74, New York, NY, USA. ACM.

\bibitem[{Zhang et~al.(2018)Zhang, Tan, and Wan}]{Zhang2018}
Jianmin Zhang, Jiwei Tan, and Xiaojun Wan. 2018.
\newblock Adapting neural single-document summarization model for abstractive
  multi-document summarization: A pilot study.
\newblock In \emph{International Conference on Natural Language Generation.}

\bibitem[{Zhang et~al.(2020)Zhang, Sun, Galley, Chen, Brockett, Gao, Gao, Liu,
  and Dolan}]{zhang2019dialogpt}
Yizhe Zhang, Siqi Sun, Michel Galley, Yen-Chun Chen, Chris Brockett, Xiang Gao,
  Jianfeng Gao, Jingjing Liu, and Bill Dolan. 2020.
\newblock Dialogpt: Large-scale generative pre-training for conversational
  response generation.
\newblock In \emph{ACL, system demonstration}.

\bibitem[{Zhu(2004)}]{zhu2004recall}
Mu~Zhu. 2004.
\newblock Recall, precision and average precision.
\newblock \emph{Department of Statistics and Actuarial Science, University of
  Waterloo, Waterloo}, 2:30.

\bibitem[{Zoph and Knight(2016)}]{zoph-knight-2016-multi}
Barret Zoph and Kevin Knight. 2016.
\newblock \href {https://doi.org/10.18653/v1/N16-1004} {Multi-source neural
  translation}.
\newblock In \emph{Proceedings of the 2016 Conference of the North {A}merican
  Chapter of the Association for Computational Linguistics: Human Language
  Technologies}, pages 30--34, San Diego, California. Association for
  Computational Linguistics.

\end{thebibliography}

% \newpage

\appendix
\section*{Appendix}

% \section{Data Pre-processing Details}
\noindent\textbf{Appendix A. Data Pre-processing Details}\\

\textbf{Data Pre-processing: } MS-MARCO Q\&A data-set \citep{DBLP:conf/nips/NguyenRSGTMD16} contains 1,010,916 questions, in which each question is associated with top-10 documents. 
Each data point contains a question and its top-10 returned documents from the Bing search engine\footnote{\url{https://www.bing.com}}.  
This question is not a target itself since not all the top-10 retrieved documents answer the question. 
However, it can give relative evaluation against a model-generated question based on the top-10 retrieved documents.
We target a broader class of problems where only document groups are available but no such group-inducing or \emph{oracle} questions.

In fact, among the top-10 retrieved documents, often one document is labeled `\textit{selected}' by human annotators to indicate that the document answers the question (\textit{true positive}), and left unknown or unlabeled for the rest of the documents, implying they may or may not answer the question (\textit{true negative} or \textit{false negative}). 
This label information is used to train the underlying generator block of their MSQG model \citep{cho_qgen_2019}. A single \textit{selected} MS-MARCO \citep{DBLP:conf/nips/NguyenRSGTMD16} document is fed into a long short-term memory-based sequence-to-sequence model to output the corresponding question. 
An example of the input \emph{selected} document is: \textit{The House of Representatives shall be composed of Members chosen every second Year by the People of the several States.... Article I, Section 2, Clause 1}, and the corresponding question is: \textit{how long is a term for a member of the house of representatives}.
We chose this dataset since the question that retrieve the top-10 documents can shed light to relative performance of our model. 

To find two 10-document sets $\dpos$ and $\dneg$ that are similar, we find a pair of questions that are semantically similar. 
However, computing pair-wise similarities among roughly 1 million questions is computationally intractable. 
Therefore, we leverage another dataset: MS-MARCO-Conversational Search\footnote{https://github.com/microsoft/MSMARCO-Conversational-Search}: an artificially constructed public dataset that simulate user search sequences.

Each data point or session is an artificial sequence of similar questions grounded on true user behavior. 
Since many similar questions are grouped together, we can reduce the search space for finding pairs of similar questions. 
Then we take pairs of high semantic similarity ($\geq0.7$) yet not a paraphrase ($\leq0.85$ following their classification criteria) using GEN-Encoder \citep{Zhang:2019:GIR:3331184.3331198} which two associated 10-document sets do not have overlaps, primarily for prototype evaluation convenience. 
For deployment models, one may choose to allow overlaps between two sets for more challenging learning.
From the two similar 10-document sets, either one is set to positive $\dpos$ or negative $\dneg$, yielding two data points for the our derived dataset.

These pre-processing steps yield 346,215 data points, each of which contains a pair of positive and negative questions, and positive and negative 10-document sets. 
Training MSCQG on the entire dataset requires processing about 7 million MARCO documents.
This is computationally intensive and takes about two days on 8 Nvidia Tesla V100 GPU cards for a single epoch. 
Therefore, for building small research prototypes and benchmarks, we will also release a subset of the data, that consists of 100K/10K/10K training, development, and evaluation data points.\newline

{\setlength{\parindent}{0cm}
\textbf{Appendix B. Data Example} 
\\

\textbf{\emph{Oracle} question for $\dpos$: }\\
number of saturn's moons\\

\textbf{\emph{Oracle} question for $\dneg$: }\\
uranus how many moons\\

\textbf{Positive Set $\dpos$ : }

\textit{
1. moons of saturn. there are 62 moons orbiting saturn. the moons of saturn vary not only in size but also in composition and shape. the largest of the moons of saturn is the aptly named titan, more than 5,000 km across and is bigger than mercury. there are 7 major moons of saturn and the rest are grouped based on the mythology from which it is taken.
}\\

\textit{
2. iapetus with a diameter of 1,470 km, it is the 3rd largest moon of saturn. it was discovered by giovanni cassini in 1671. it has a distinct feature of having a bright and dark hemisphere. dione the 4th largest moon of saturn named after a vague character in greek mythology.
}\\

\textit{
3. titan is the largest of saturn's moons and the first to be discovered. titan is the only moon in the solar system known to have a significant atmosphere. nitrogen and methane extend around the moon 10 times as far into space as earth's atmosphere, sometimes falling to the surface in the form of methane rain. }\\

\textit{
4. saturn has at least 150 moons and moonlets, 53 of which have formal names. titan, the largest, comprises more than 90\% of the mass in orbit around saturn, including the rings. saturn's second-largest moon, rhea, may have a tenuous ring system of its own, along with a tenuous atmosphere. }\\

\textit{
5. their journeys around the ringed planet average from half an earth day to just over four earth years. saturn's moons formed early in the history of the solar system. one of the moons, titan, makes up 96 percent of the mass orbiting the planet. scientists think that the system may have originally housed two such moons, but the second broke up, creating the debris that formed the rings and smaller, inner moons.} \\

\textit{
6. saturn has a prominent ring system that consists of nine continuous main rings and three discontinuous arcs and that is composed mostly of ice particles with a smaller amount of rocky debris and dust. sixty-two moons are known to orbit saturn, of which fifty-three are officially named. }\\

\textit{
7. sixteen of the moons are tidally locked, with one face permanently turned toward saturn. the first moon was discovered in 1655. over the next 200 years, the other seven major satellites were spotted. by 1997, astronomers on earth had found 18 moons in orbit around the planet. }\\

\textit{
8. saturn is the sixth planet from the sun and the second-largest in the solar system, after jupiter. it is a gas giant with an average radius about nine times that of earth. although only one-eighth the average density of earth, with its larger volume saturn is just over 95 times more massive. }\\

\textit{
9. this temporary name usually consists of the year of discovery and a number indicating the order of discovery in that year. in the case of saturn's moons, these provisory names follow the format s/2005-s1, s/2005-s2 etc. the first s (before the slash) is for saturn. the second s (after the dash) is for satellite.}\\

\textit{
10. this does not include the hundreds of moonlets comprising the rings. titan, saturn's largest moon, and the second-largest in the solar system, is larger than the planet mercury, although less massive, and is the only moon in the solar system to have a substantial atmosphere.}\\

\textbf{Negative Set $\dneg$: }

\textit{
11. uranus has 27 moons that we know of. five of the moons are large and the rest are much smaller. the five large moons are called miranda, ariel, umbriel, titania, and oberon. titania is the largest moon of uranus and it is covered with small craters, a few large craters, and very rough rocks. ariel is the brightest moon of uranus and has canyons and valleys as well as a lot of craters. umbriel is very dark. }\\

\textit{
12. uranus can't seem to catch a break these days. besides spinning on its side like the drunkard of the solar system and being the butt of everyone's jokes, new research suggests several of its tiny moons will collide in a million years. uranus can't seem to catch a break these days. }\\

\textit{
13. the gas giant uranus is the third largest planet in our solar system, has many moons, a ring system, and composed of gases and ices. universe today space and astronomy news login }\\

\textit{
14. the researchers used cressida's mass and orbit to determine its possible doom. since uranus' 27 moons are tightly packed together, the team posits that in a million years, cressida will likely have a deadly encounter with one of its neighboring moons, called desdemona. previous research and simulations suggest cupid and belinda will also probably smack into each other some time between 1,000 and 10 million years from now. }\\

\textit{
15. puck, at 162 km, is the largest of the inner moons of uranus and the only one imaged by voyager 2 in any detail while puck and mab are the two outermost inner satellites of uranus. all inner moons are dark objects.
}\\

\textit{
16. uranus, which takes its name from the greek god of the sky, is a gas giant and the seventh planet from our sun. it is also the third largest planet in our solar system, ranking behind jupiter and saturn. like its fellow gas giants, it has many moons, a ring system, and is primarily composed of gases that are believed to surround a solid core. }\\

\textit{
17. in 1986, the voyager 2 spacecraft hit the jackpot while studying uranus and discovered 10 other moons, including desdemona and cressida. since then, hubble observations have helped bring that number up to 27 for now.
}\\

\textit{
18. at an average distance of 3 billion km from the sun, it takes uranus roughly 84 years (or 30,687 days) to complete a single orbit of the sun. 1 the rotational period of the interior of uranus is 17 hours, 14 minutes. as with all giant planets, its upper atmosphere experiences strong winds in the direction of rotation. }\\

\textit{
19. uranus' size, mass and orbit: with a mean radius of approximately 25,360 km, a volume of 6.833—10^13 km3, and a mass of 8.68 — 10^25 kg, uranus is approximately 4 times the sizes of earth and 63 times its volume.
}\\

\textit{
20. uranus has 27 known satellites, which are divided into the categories of larger moons, inner moons, and irregular moons (similar to other gas giants). the largest moons of uranus are, in order of size, miranda, ariel, umbriel, oberon and titania. }\\

}

% \section{Retrieval Baselines}
\noindent\textbf{Appendix C. Retrieval Baselines}\\

\textbf{Top-TFIDF@K} and \textbf{Top-Frequent@K}

The retrieval baselines are designed to give a relative insight into the performance between MSQG in \citet{cho_qgen_2019} and our novel coordinator model. 
We use Lucene to retrieve \emph{questions} instead of documents from a corpus composed of the 1,010,916 MS-MARCO questions.
The retrieved questions from Top-TFIDF@K and Top-Frequent@K baselines are evaluated in the same manner as the generated ones.

For the intersection to be non-empty, $k$ should be sufficiently large.
However, even for $k=1000$, there were no intersecting subset questions for almost all cases. 
Therefore, we relax the intersection among all 10 retrieved sets, into finding the most frequently occurring question among the 10 top-$k$ retrieved sets. 
$k=100$ was an appropriate value that is not too large to retrieve remotely relevant questions, and not too small to yield vastly different retrieval sets. 
If there are multiple questions with the same count, we randomly choose one.

\begin{algorithm}[H]
\caption{Top-TFIDF@K}
\label{pseudoPSO}
\begin{algorithmic}[0]
\State \textbf{Input: } $\dpos, \textbf{Corpus }\mathbb{C}$
\State For each $d \in \dpos$, retrieve top-K questions in $\mathbb{C}$;
\State Using all unique questions $\mathbb{Q}$, compute TF-IDF;
\State Let $\Psi$ be the TF-IDF transform operator;
\State $q^*=\argmax\limits_{q\in\mathbb{Q}} \sum\limits_{d\in\dpos} \cossim\left(\Psi_q,\Psi_d\right)$;
\State \textbf{Output: }$q^*$
\end{algorithmic}
\end{algorithm}

\begin{algorithm}[H]
\caption{Top-Frequent@K}
\label{pseudoPSO}
\begin{algorithmic}[0]
\State \textbf{Input: } $\dpos, \textbf{Corpus }\mathbb{C}$
\State For each $d \in \dpos$, retrieve top-K questions in $\mathbb{C}$;
\State Let $\mathbb{S}_d$ be the retrieved set for each $d$;
\State $q^*=\argmax\limits_{q\in\mathbb{Q}} \sum\limits_{d\in\dpos} \mathbbm{1}_{q \in \mathbb{S}_d}$;
\State \textbf{Output: }$q^*$
\end{algorithmic}
\end{algorithm}

% \section{Experiment Configurations}
\noindent\textbf{Appendix D. Experiment Configurations}\\

\textbf{Document-specific GPT-2 Generator: } 
From each document $i$, the generator yields its final layer hidden state $h_i \in \mathbb{R}^{H}$ ($H=768$ is the hidden dimension) and a document-specific discrete output distribution $\pi_i \in \mathbb{R}^{V}$ ($V=50257$ is the vocabulary dimension) from the learned language model head. 

\textbf{Coordinator:} The input size is 20 with the dimensionality of the embeddings and hidden states as 768. 
The number of recurrent layers is 2, with 4 attention heads in each layer. 
The epsilon value used in the layer normalization is set to $1\mathrm{e}{-5}$. 
The number of cluster embeddings is 2 (positive or negative). 
The standard deviation of the truncated normal initializer for weight matrices is 0.02.
$\lambda_1,\lambda_2,\lambda_3=1.0,100.0,0.1$.
We performed coarse hyper-parameter search for equidistant values in log-scale for all $\lambda_1,\lambda_2,\lambda_3$, and use the best configuration.
Maximum generation length is 20 tokens. 
We use the BERT \citep{devlin2018bert} version of Adam optimizer \citep{Kingma2014AdamAM} with weight decay of 0.01 and learning rate of $1\mathrm{e}{-5}$.

We trained the coordinator model by maximizing \textit{Precision@10} with \emph{oracle} questions as the policy gradient baseline. It is reasonable to weigh the documents unevenly because often times not all the top-10 retrieved documents from the Bing search engine share the same content.
Thus, we leave to the model to learn the optimal attention weights among positive and negative sets that produce a more grounded question.
Additional experiment results using a different baseline - self-critic \citep{rennie2017self} - is shown in Table~\ref{additional_overall_retrieval}.
This shows that our proposed model framework is effective even with any of the two policy gradient baselines.
Conceptually, the coordinator model would generate a question that can better retrieve the documents from the positive set, aided by the negative set. \\

% \section{Human Evaluation Details}
\noindent\textbf{Appendix E. Human Evaluation Details}\\

We performed two human evaluations: In the first experiment, we showed judges one randomly selected positive document, which is about 300 words long, followed by a pair of questions from the three sources. Judges were asked to evaluate which one of the two questions is preferred based on four criteria. For each pair of three sources, we evaluated 200 same random samples for each judge (or 600 samples for the 3 judges), totalling 1,800 samples.
 
In the second experiment, human annotators evaluated contrastive ability from 1,800 samples of one question, followed by two documents each from the positive and negative sets. Note that our model is trained to generate questions, accounting for the negative set.

The results were averaged across all samples and judges. 

For computing the Pearson correlation between the ranker and human evaluation results, we map \textit{Option A preferred} $\rightarrow$ 0, \textit{Same} $\rightarrow$ 0.5, \textit{Option B preferred} $\rightarrow$ 1, accounting for the random assignments between \textit{A} and \textit{B}. This projection ensures that image of two metrics are the same (between 0 and 1). 
Then we compute the correlation value between two results.

\begin{table*}[h!]
\resizebox{\textwidth}{!}{%
\begin{tabular}{l|cccc|ccccc}
 & \multicolumn{4}{c|}{\textbf{Out-Sample IR}} & \multicolumn{5}{c}{\textbf{Search-Engine Augmented IR}} \\ \hline
\textbf{Model} & mAP & RPrec & MRR (=MRR@10) & nDCG & mAP & RPrec & MRR & MRR@10 & nDCG \\ \hline
Top-TFIDF @100 & 0.416 & 0.533 & 0.696 & 0.545 & 0.113 & 0.0588 & 0.0260 & 0.0050 & 0.181 \\
Top-Frequent @100 & 0.680 & 0.742 & 0.921 & 0.779 & 0.171 & 0.129 & 0.0404 & 0.0119 & 0.204 \\ \hline
MSQG (Cho et al. '19) & - & - & - & - & - & - & 0.0704 & 0.0441 & 0.234 \\ \hline
$\text{MSQG}_{GPT2}$ & 0.713 & 0.763 & 0.945 & 0.804 & 0.245 & 0.217 & 0.0714 & 0.0400 & 0.240 \\
$\text{MSCQG}_{SCR}$ & 0.751 & 0.790 & 0.974 & 0.836 & 0.258 & 0.234 & 0.0745 & 0.0420 & 0.245 \\

\hline
$\text{MSCQG}_{PG+SCR+H}^{\textnormal{self-critic,null-neg}}$ & 0.714 & 0.764 & 0.945 & 0.805 & 0.247 & 0.220 & 0.0724 & 0.0407 & 0.241  \\
$\text{MSCQG}_{PG}^{\textnormal{self-critic}}$ & \textbf{0.762} & \textbf{0.798} & \textbf{0.982} & \textbf{0.845} & 0.259 & 0.237 & 0.0746 & 0.0420 & 0.244 \\
$\text{MSCQG}_{PG+SCR}^{\textnormal{self-critic}}$ & 0.760 & 0.797 & 0.977 & 0.843 & 0.260 & 0.236 & 0.0744 & 0.0416 & 0.245 \\
$\text{MSCQG}_{PG+SCR+H}^{\textnormal{self-critic}}$ & 0.760 & 0.797 & 0.977 & 0.843 & \textbf{0.262} & \textbf{0.238} & \textbf{0.0771} & \textbf{0.0444} & \textbf{0.247} \\ \hline
$\text{MSCQG}_{PG+SCR+H}^{\textnormal{orcl-critic,null-neg}}$ & 0.717 & 0.766 & 0.950 & 0.808 & 0.246 & 0.220 & 0.0722 & 0.0404 & 0.241 \\
$\text{MSCQG}_{PG}^{\textnormal{orcl-critic}}$ & 0.753 & 0.791 & 0.978 & 0.838 & 0.256 & 0.232 & 0.0742 & 0.0421 & 0.244 \\
$\text{MSCQG}_{PG+SCR}^{\textnormal{orcl-critic}}$ & \textbf{0.767} & \textbf{0.803} & \textbf{0.981} & \textbf{0.849} & \textbf{0.265} & \textbf{0.242} & 0.0748 & 0.0420 & 0.245 \\
$\text{MSCQG}_{PG+SCR+H}^{\textnormal{orcl-critic}}$ & 0.765 & 0.800 & 0.976 & 0.847 & 0.262 & 0.239 & \textbf{0.0759} & \textbf{0.0434} & \textbf{0.246}\\\hline
\emph{Oracle} Questions for $\dpos$& 0.759 & 0.797 & 0.976 & 0.842 & 0.292 & 0.273 & 0.0846 & 0.0495 & 0.256 \\ \hline
\end{tabular}%
}
\caption{Additional retrieval performance using self-critic \citep{rennie2017self} baseline in the policy gradient, applicable to datasets with no \emph{oracle} questions.
It shows that our framework is also effective using a different baseline.
The superscript \emph{null-neg} denotes models that do not use negative attentions when generating questions. 
This shows the importance of the negative set in promoting specificity in the generated question.
It further corroborates that the non-uniform weighted-sum scheme among $\dpos$ improves performance because not all documents in $\dpos$ revolve around the same topic, and the model learns to address this nature of the dataset through unequal weights and generate a more representative question.
}
\label{additional_overall_retrieval}
\end{table*}

\end{document}